\documentclass[10pt]{article}

\usepackage[accepted]{tmlr}

\usepackage{amsmath}
\usepackage{amssymb}
\usepackage{amsthm}

\usepackage{float}
\usepackage{graphicx}
\usepackage{booktabs}
\usepackage{xcolor}
\usepackage{colortbl}
\usepackage{caption}
\usepackage[linesnumbered,ruled,vlined]{algorithm2e}
\let\origincludegraphics\includegraphics \renewcommand{\includegraphics}[2][]{\scalebox{0.97}{\origincludegraphics[#1]{#2}}}
\makeatletter
\def\section{\@startsection{section}{1}{\z@}{-2.0ex plus -0.5ex minus -.2ex}{0.2ex plus 0.1ex}{\large\bf\raggedright\sffamily}}
\def\subsection{\@startsection{subsection}{2}{\z@}{-0.9ex plus -0.3ex minus -.1ex}{0.2ex plus 0.1ex}{\normalsize\bf\raggedright\sffamily}}
\def\subsubsection{\@startsection{subsubsection}{3}{\z@}{-0.75ex plus -0.3ex minus -.1ex}{0.2ex plus 0.1ex}{\normalsize\bf\raggedright\sffamily}}
\makeatother

\usepackage{hyperref}
\usepackage{url}
\usepackage{cleveref}

\crefname{section}{Section}{Sections}
\Crefname{section}{Section}{Sections}
\crefname{subsection}{Section}{Sections}
\Crefname{subsection}{Section}{Sections}
\crefname{figure}{Figure}{Figures}
\Crefname{figure}{Figure}{Figures}
\crefname{table}{Table}{Tables}
\Crefname{table}{Table}{Tables}
\crefname{equation}{Eq.}{Eqs.}
\Crefname{equation}{Equation}{Equations}
\crefname{algorithm}{Algorithm}{Algorithms}
\Crefname{algorithm}{Algorithm}{Algorithms}
\crefname{appendix}{Appendix}{Appendices}
\Crefname{appendix}{Appendix}{Appendices}

\newlength{\itemizelength}

\def\month{09}
\def\year{2026}
\def\openreview{\url{https://openreview.net/forum?id=hlNAYdhUi6}}

\title{Self-Supervised Representation Learning\\
    as Mutual Information Maximization}

\author{\name Akhlaqur Rahman Sabby \email sabby@dal.ca \\
        \addr Dalhousie University
        \AND
        \name Yi Sui \email amy@layer6.ai \\
        \addr Layer 6 AI
        \AND
        \name Tongzi Wu \email tongzi@layer6.ai \\
        \addr Layer 6 AI
        \AND
        \name Jesse C.\ Cresswell \email jesse@layer6.ai \\
        \addr Layer 6 AI
        \AND
        \name Ga Wu \email ga.wu@dal.ca \\
        \addr Dalhousie University}

\begin{document}
\maketitle

\begin{abstract}
Self-supervised representation learning (SSRL) has demonstrated remarkable empirical success, yet its underlying principles remain insufficiently understood. While recent works attempt to unify SSRL methods by examining their information-theoretic objectives or summarizing their heuristics for preventing representation collapse, architectural elements like predictor networks, stop-gradient operations, and statistical regularizers are often viewed as empirically motivated additions. In this paper, we adopt a first-principles approach and investigate whether the learning objective of an SSRL algorithm dictates its possible optimization strategies and model design choices. In particular, by starting from a variational mutual information (MI) lower bound, we derive two training paradigms, namely Self-Distillation MI (SDMI) and Joint MI (JMI), each imposing distinct structural constraints and covering a set of existing SSRL algorithms. SDMI relies on alternating optimization, in which stop-gradient operations serve as a principled mechanism for realizing the alternating updates. In contrast, JMI admits joint optimization through symmetric architectures without such components. Under the proposed formulation, predictor networks in SDMI and statistical regularizers in JMI emerge as tractable surrogates for the MI objective. We show that many existing SSRL methods are specific instances or approximations of these two paradigms. This paper provides a theoretical explanation for the choices of different architectural components of existing SSRL methods, going beyond heuristic conveniences.\footnote{Code: \url{https://github.com/AkhlaqurRahmanSabby/SSRL-as-MI-Maximization}}
\end{abstract}

\section{Introduction}
\label{sec:introduction}
SSRL has demonstrated remarkable success at learning useful features from unlabeled data, achieving competitive performance with supervised approaches across a wide range of tasks \citep{lecun2015deep, bengio2013representation, balestriero2023cookbook}. Conventionally, SSRL algorithms are divided into two categories according to their training objectives, namely contrastive methods and non-contrastive methods. Contrastive methods \citep{oord2018representation, tian2020contrastive, chen2020simple, he2020momentum, chen2020improved, chen2021empirical} train a representation model by aligning the representations of two augmentations of the same input while explicitly pushing apart those of augmentations of different inputs. Non-contrastive methods \citep{grill2020bootstrap, chen2020exploring, caron2021emerging, zbontar2021barlow, bardes2021vicreg, sui2023self}, by contrast, challenge the necessity of negative samples and propose alternative mechanisms, such as momentum encoders or stop-gradient operations, to prevent representational collapse.

Many recent studies attempt to unify these two categories of SSRL methods under common theoretical frameworks, often through shared information-theoretic principles. \citet{liu2022self} interpret various SSRL methods as low-order approximations of a unified maximum entropy principle. \citet{zbontar2021barlow} motivate the Barlow Twins objective via Information Bottleneck theory \citep{tishby2000information, tishby2015deep}, which \citet{tsai2021note} later relate to a kernel-based MI measure. \citet{shwartz2023information} connect VICReg's penalties to MI bounds, and \citet{jha2024common} propose a unifying framework that explains collapse avoidance by minimizing a global mean while preserving augmentation-level variation. Despite their insights, prior work offers little clarity on whether training strategies like self-distillation or variance--covariance control are necessary consequences of the learning objective, principled approximations to it, or independent design choices, leaving an important gap in our theoretical understanding of these methods.

In this work, we close this gap by returning to first principles, analyzing existing SSRL algorithms through the lens of MI maximization, a perspective implicitly or explicitly invoked by many self-supervised learning methods.
Starting from a variational lower bound on MI, particularly the Donsker--Varadhan (DV) bound, we show that this bound naturally leads to two optimization paradigms in the context of SSRL. The first, Self-Distillation MI (SDMI), uses EM-style alternating updates with stop-gradient operations (e.g., SimSiam, BYOL, MoCo). The second, Joint MI (JMI), supports joint optimization via a single gradient step per batch (e.g., SimCLR, Barlow Twins, VICReg). We further argue that this taxonomy is more theoretically principled than the traditional contrastive vs.\ non-contrastive distinction, as it is derived directly from the DV bound rather than from heuristic descriptions of training behavior. We then distill SDMI and JMI into canonical algorithmic forms, and demonstrate that these forms behave similarly to existing SSRL methods in their respective paradigms and remain competitive with them on downstream tasks.

In summary, we make the following contributions.
\begin{enumerate}
    \item We formulate a general MI maximization perspective under the DV bound, showing that existing SSRL methods implicitly follow one of two optimization paradigms, namely Self-Distillation MI (SDMI) or Joint MI (JMI).
    \item We show that design elements such as stop-gradients, predictor networks, and statistical regularizers (typically introduced as empirical fixes) arise as principled consequences of optimizing an MI lower bound under one of the two paradigms, providing a formal explanation for these common design choices.
    \item We show that many well-known SSRL approaches (e.g., SimCLR, BYOL, SimSiam) can be mapped directly to our two paradigms. This helps unify a broad set of SSRL methods under a shared theoretical lens and offers guidance for future method design.
\end{enumerate}

\section{Related Work and Preliminaries}
\label{sec:preliminaries}
We begin by reviewing recent attempts to unify the growing landscape of SSRL methods under shared theoretical principles. We first summarize key unification efforts based on objective design and collapse-prevention mechanisms, highlighting their contributions and limitations. We then present MI as a foundational concept and starting point for our analysis, revisiting its definition and variational lower bounds, with a focus on the DV bound, whose tightness and decomposition are central to our work.

\subsection{Unification approaches in SSRL}

A growing body of work \citep{zbontar2021barlow, liu2022self, tsai2021note, shwartz2023information, jha2024common, tan2023information} suggests that an information-theoretic lens can help unify seemingly disparate SSRL methods. Many existing methods, particularly contrastive approaches, can be explicitly framed as maximizing MI between representations of different augmented views \citep{oord2018representation, he2020momentum, chen2020improved, chen2021empirical, poole2019variational}.

Building on this, several works link specific SSRL losses to MI estimation. The Information Bottleneck perspective \citep{tishby2000information, tishby2015deep} is applied to Barlow Twins \citep{zbontar2021barlow}, and \citet{tsai2021note} show that the Barlow Twins objective is equivalent to maximizing a Hilbert--Schmidt Independence Criterion \citep{gretton2005measuring}, a kernelized dependence measure related to MI. \citet{bardes2021vicreg} introduce VICReg's variance and covariance penalties, and \citet{shwartz2023information} later provide an information-theoretic analysis linking these penalties to MI bounds.

Another prominent unification direction is offered by \citet{liu2022self}, who propose a Maximum Entropy Coding (MEC) framework that treats representation learning as an entropy maximization problem, showing that many existing SSRL methods can be interpreted as low-order Taylor approximations of a single entropy-based objective. Complementing this view, \citet{jha2024common} analyze the collapse avoidance mechanisms that ensure stability in SSRL, arguing that, despite architectural and algorithmic differences, most methods implicitly minimize the global average of learned representations while preserving sample-level variability. From a structural perspective, \citet{nguyen2023lvm} generalize contrastive and instance-based objectives within a second-order graph matching formulation, where edge affinities encode the relational structure of embeddings across two augmented views of a batch.

While these approaches provide valuable unifying perspectives, they operate at the level of the learning objective, showing that disparate SSRL losses estimate or bound a common quantity such as entropy \citep{liu2022self}, a kernelized dependence measure \citep{tsai2021note}, or MI itself \citep{shwartz2023information}. A separate line of work studies individual components, taking the predictor and stop-gradient as given and analyzing how they prevent collapse \citep{tian2021understanding, zhang2022does, wen2022mechanism}. Both leave open a complementary question at the level of optimization: whether components such as stop-gradient operations, predictor networks, and statistical regularizers are necessary consequences of the learning objective, principled approximations to it, or independent design choices. We address this question starting from the DV bound, whose decomposition into a joint (alignment) term and a marginal (log-partition) term (\cref{sec:section_2.2}) separates the two forces any SSRL method that learns from augmented views must balance, exposing both the block-coordinate structure behind self-distillation (\cref{sec:section_3.1}) and the marginal-term role that predictors and regularizers fill.

\subsection{Mutual information and its variational bounds in SSRL}
\label{sec:section_2.2}

In SSRL, MI is often defined between representations \(Z_A\) and \(Z_B\) of two augmented views \(X_A\) and \(X_B\) of an input \(X\) in the form \(I(Z_A; Z_B) = D_{\mathrm{KL}} \left[ p(z_A, z_B) \,\|\, p(z_A)p(z_B) \right]\). Maximizing MI with respect to the encoding function \(Z_A = f_{\theta}(X_A)\) defines a valid pretext task for learning representations that can transfer to various downstream applications. However, direct optimization of MI is intractable since the underlying data distribution \(P(X)\) is unknown, motivating the use of variational bounds in practice. Common variational bounds include InfoNCE \citep{oord2018representation, poole2019variational}, Barber--Agakov \citep{barber2004algorithm}, TUBA \citep{poole2019variational}, NWJ \citep{nguyen2010estimating}, JSD \citep{hjelm2018learning} and DV \citep{belghazi2018mine}. Each of these alternatives introduces different trade-offs between tightness, stability, and optimization feasibility.

We use the Donsker--Varadhan (DV) bound to guide our analysis in this paper, as it offers (1) a direct connection to MI via KL divergence, (2) a natural variational decomposition that facilitates block-coordinate ascent, and (3) provable tightness relative to $f$-divergence-based alternatives for any fixed function class \citep{belghazi2018mine}.

\textbf{Donsker--Varadhan bound}\quad Over a sufficiently rich class of functions \(\mathcal{F}\), the DV bound decomposes MI as
\begin{equation}
    \label{eq:dv_bound}
    I(Z_A; Z_B) \ge I_{\mathrm{DV}}(Z_A; Z_B) {=} \sup_{T \in \mathcal{F}} \bigg\{ \underbrace{\mathbb{E}_{p(z_A, z_B)}[T(z_A, z_B)]}_{\text{Joint term}} {-} \underbrace{\log \mathbb{E}_{p(z_A)p(z_B)}\left [e^{T(z_A, z_B)} \right]}_{\text{Marginal term}} \bigg\},
\end{equation}
where \(\mathcal{F} \subseteq \{\, f : \mathcal{Z}_A \times \mathcal{Z}_B \to \mathbb{R} \,\}\), and \(T \in \mathcal{F}\) is a scoring function that assigns high values to joint pairs \((z_A, z_B) \sim p(z_A, z_B)\) and low values to marginal pairs \((z_A, z_B) \sim p(z_A)p(z_B)\).

\section{A Unified View of SSRL as MI Maximization}
\label{sec:method}
In this section, we first revisit the DV lower bound on MI from an optimization perspective. This gives rise to two natural optimization paradigms in SSRL, namely Self-Distillation MI (SDMI) and Joint MI (JMI). We then analytically show how a wide range of SSRL methods can be categorized under these paradigms.

\subsection{Block-coordinate ascent via DV bound}
\label{sec:section_3.1}

Let representations $Z_A$ and $Z_B$ come from two different encoding functions $f_\theta$ and $g_{\xi}$ with a fixed scoring function $T$ drawn from the function class $\mathcal{F}$. We note that the DV bound in \cref{eq:dv_bound} provides a useful formulation for optimization, since exactly maximizing the bound with respect to $\theta$ while holding $\xi$ fixed (and vice versa) guarantees a non-decreasing improvement of the objective. As a result, alternating updates over the encoders for $Z_A$ and $Z_B$ constitute valid block-coordinate ascent steps. Specifically, we formalize the improvement in the following proposition.

\paragraph{Proposition}
Let the DV-bound objective be given by
\begin{equation}
\label{eq: general}
   \mathcal{L}(\theta, \xi) = J(\theta; \xi) - M(\theta; \xi),
\end{equation}
where $\mathcal{L}(\theta, \xi)$ is the DV bound, $J(\theta; \xi)$ is the joint term, and $M(\theta; \xi)$ is the marginal term from \cref{eq:dv_bound}.

Assume that (1) for fixed $\xi$, $J(\cdot; \xi)$ is concave in $\theta$; (2) the marginal term $M(\cdot; \xi)$ is smooth and satisfies $\Vert\nabla_\theta M(\theta; \xi)\Vert \le \varepsilon$; and (3) the same conditions hold symmetrically for updates over $\xi$. Then alternating gradient steps over $\theta$ and $\xi$ yield approximate monotonic improvement in $\mathcal{L}(\theta, \xi)$:
\[
\mathcal{L}(\theta^{(k+1)}, \xi^{(k)}) \ge \mathcal{L}(\theta^{(k)}, \xi^{(k)}) - \mathcal{O}(\varepsilon),
\quad
\mathcal{L}(\theta^{(k+1)}, \xi^{(k+1)}) \ge \mathcal{L}(\theta^{(k+1)}, \xi^{(k)}) - \mathcal{O}(\varepsilon).
\]

See \Cref{app:block_coordinate} for our proof.

\textbf{Scope and interpretation}\quad These assumptions are idealized. For deep networks, concavity in the encoder parameters (Assumptions 1 and 3) does not hold globally, though it may hold in a neighborhood of a local optimum, so the proposition offers an idealized justification of alternating updates rather than a convergence guarantee; with respect to the scoring function, however, the structure is exact, since the joint term is affine and the log-partition term is convex in $T$. Assumption 2 requires the marginal term to vary slowly with the encoder parameters, which batch normalization, slowly drifting EMA targets, and large batches all encourage. Strict monotonic improvement holds only when $\varepsilon = 0$, and for $\varepsilon > 0$ the proposition instead bounds how much the objective can decrease in each step. When $\Vert\nabla_\theta M\Vert$ is small, the joint term drives every update, and the proposition then states that alternating updates on the joint term alone approximately maximize the full DV bound as long as the marginal statistics remain stable; this is the regime in which self-distillation methods operate.

When sharing parameters $\theta = \xi$, the maximization objective in \cref{eq: general} can be jointly optimized via standard gradient ascent with the guarantee of monotonic improvement, provided that the full objective $\mathcal{L}(\theta)$ is concave. In the DV bound (\cref{eq:dv_bound}), the joint term is affine and the marginal term is convex with respect to the scoring function $T$ (as a log-partition function), providing a principled motivation for joint gradient ascent when $\theta = \xi$.

The same block-coordinate argument extends to other variational MI bounds with the joint-minus-marginal structure, such as InfoNCE and JSD; see \cref{app:other_variational_bounds}.

As such, there are two valid optimization paradigms to maximize MI: alternating updates across encoder branches or joint updates over shared parameters. We name the two paradigms Self-Distillation MI (SDMI) and Joint MI (JMI), respectively.

\subsection{Self-Distillation Mutual Information (SDMI)}

SSRL methods in the SDMI paradigm rely on an EM-style alternating update schedule between two encoder branches and on a mechanism for maximizing MI between augmented views. The alternating updates are enabled through a stop-gradient operator, which breaks the gradient flow from one branch to the other, making it possible to treat one encoder as fixed while updating the other, mimicking a block-coordinate ascent on the DV bound. Typically, these methods use an online encoder that receives direct gradient updates and a target (or momentum) encoder that is updated via an exponential moving average (EMA) of the online encoder's parameters. While some existing SDMI methods such as SimSiam and BYOL do not explicitly optimize a variational MI bound, we show that their alternating update structure, enabled by stop-gradients and architectural asymmetry, can be derived as a principled optimization strategy for DV-bound maximization. This provides a theoretical justification for design choices that were previously considered heuristic.

\paragraph{Block-coordinate interpretation of SDMI}
\label{sec:EM_style_MI_maximization}

To formalize SDMI as an EM-style block-coordinate ascent procedure, we consider batches of two augmented views $X_1 = \{x_1^i\}_{i=1}^N$ and $X_2 = \{x_2^i\}_{i=1}^N$, where each $x_1^i, x_2^i$ is sampled from a stochastic augmentation $\mathcal{A}(x)$ applied to an input $x \sim P(x)$. We denote the batch size by $N$ and use two encoders $f_\theta$ and $g_\xi$.

\textbf{E-Step:}
At iteration $k$, we define the MI between the representations produced by the encoders $f_\theta$ and $g_\xi$ as
\begin{equation}
    I^{(k)} = I\big(f_{\theta^{(k)}}(X_1),\; g_{\xi^{(k)}}(X_2)\big).
\end{equation}
We update the $f_\theta$ encoder by maximizing the objective under a stop-gradient (SG) on the $g_\xi$ encoder:
\begin{equation}
\label{E_step_update}
    \theta^{(k+1)} = \arg\max_{\theta}\; I\big(f_\theta(X_1),\; \mathrm{SG}(g_{\xi^{(k)}}(X_2))\big),
\end{equation}
which guarantees $I\big(f_{\theta^{(k+1)}}(X_1),\; g_{\xi^{(k)}}(X_2)\big) \ge I\big(f_{\theta^{(k)}}(X_1),\; g_{\xi^{(k)}}(X_2)\big)$.

\textbf{M-Step:}
Using the updated $f_\theta$ encoder, we update the $g_\xi$ encoder with a stop-gradient on $f_\theta$,
\begin{equation}
\label{M_step_update}
    \xi^{(k+1)} = \arg\max_{\xi}\; I\big(\mathrm{SG}(f_{\theta^{(k+1)}}(X_1)),\; g_\xi(X_2)\big),
\end{equation}
ensuring $I\big(f_{\theta^{(k+1)}}(X_1),\; g_{\xi^{(k+1)}}(X_2)\big) \ge I\big(f_{\theta^{(k+1)}}(X_1),\; g_{\xi^{(k)}}(X_2)\big)$.

\textbf{Monotonic Improvement:}
Together, these steps guarantee overall monotonic improvement:
\begin{equation}
    I\big(f_{\theta^{(k+1)}}(X_1),\; g_{\xi^{(k+1)}}(X_2)\big)
    \ge I\big(f_{\theta^{(k+1)}}(X_1),\; g_{\xi^{(k)}}(X_2)\big)
    \ge I\big(f_{\theta^{(k)}}(X_1),\; g_{\xi^{(k)}}(X_2)\big).
\end{equation}

\subsubsection{Examples of SDMI methods}

\paragraph{SimSiam and BYOL}
\label{sec:byol_and_simsiam}

Non-contrastive methods, such as BYOL \citep{grill2020bootstrap} and SimSiam \citep{chen2020exploring}, fall under the SDMI paradigm. These methods train an online encoder $f_{\theta}$, together with a lightweight predictor $h_{\phi}$, to align transformed representations with those of a target encoder $g_{\xi}$. From the SDMI viewpoint, both methods approximate a two-step EM-style optimization in a relaxed, implicit form, as described below.

\begin{enumerate}
\setlength{\leftskip}{\itemizelength}
  \item \textbf{E-step:} Both methods update the online encoder to maximize MI by minimizing the following negative cosine similarity loss
    \begin{equation}
        \label{eq:simsiam_byol_objective}
        \min_{\theta,\phi}\; -\,\mathbb{E}_{p(x_1,x_2)}\bigl[T_\mathrm{cos}\bigl(h_\phi(f_\theta(x_1)),\,g_\xi(x_2)\bigr)\bigr],
    \end{equation}
    where $T_\mathrm{cos}$ denotes a cosine similarity scoring function. This loss can be viewed as an instantiation of the DV bound with cosine similarity, which we refer to as $I_{\text{cos-DV}}$. However, these methods omit the explicit marginal term present in the full bound (see \cref{eq:cos_dv_bound}), relying instead on their predictor dynamics to discourage collapse.

  \item \textbf{M-step (Implicit):} Immediately after the E-step, SimSiam resets the target encoder with the new online weights and freezes it for the next E-step:
    \begin{equation}
      \label{eq:simsiam_m_step}
      g_\text{new} = \mathrm{SG}(f_\theta).
    \end{equation}
    BYOL, on the other hand, uses an EMA of $\theta$:
    \begin{equation}
      \label{eq:byol_m_step}
      \xi \leftarrow \tau \xi + (1 - \tau) \theta.
    \end{equation}
    While these methods differ from SDMI's explicit coordinate ascent step on the $g_\xi$ encoder, they preserve the underlying principle of alternating optimization, though in an implicit form.
\end{enumerate}

Our interpretation aligns with the hypothesis of \citet{chen2020exploring} that SimSiam's stop-gradient induces EM-like alternating updates between online and frozen branches. While they suggest that the predictor approximates an expectation over augmentations, \citet{zhang2022does} refute this, showing instead that it induces de-centering and de-correlation gradients that stabilize training and promote feature diversity. Within our SDMI framework, we reinterpret these effects as implicitly approximating the marginal term of the DV bound. \Cref{sec:predictor_ablation} provides further analysis and empirical evidence supporting this interpretation.

\textbf{MoCo}\quad MoCo \citep{he2020momentum, chen2020improved, chen2021empirical}, a contrastive learning method, also fits naturally within the SDMI paradigm. It performs EM-style alternating updates between an online encoder and a momentum encoder, while directly optimizing the InfoNCE lower bound on MI. Its momentum encoder plays a similar functional role and is updated via EMA, like the target encoder in BYOL. Early versions (MoCo-v1 \citep{he2020momentum}, v2 \citep{chen2020improved}) already achieve strong performance without predictor networks, and although MoCo-v3 \citep{chen2021empirical} introduces a predictor, it yields only marginal performance gains $(\sim 1\%)$, underscoring that with direct MI maximization, predictors are auxiliary.

This illustrates how the SDMI framework unifies both traditional contrastive and non-contrastive methods under a shared lens of MI maximization with alternating encoder updates.

\subsection{Joint Mutual Information (JMI)}

Unlike SDMI, JMI methods use a single encoder $f_\theta$ to produce representations for both augmented views, enabling joint gradient updates to maximize MI. Joint optimization is achieved either by directly optimizing an explicit MI objective or by incorporating surrogate regularization terms that penalize statistical properties, such as variance, covariance, or feature redundancy, to approximate the marginal log-partition term in \cref{eq:dv_bound}. A general JMI objective written as a loss function takes the form
\begin{equation}
\label{jmi_objective}
    \mathcal{L}_{\mathrm{JMI}} = -\mathbb{E}_{p(x_1, x_2)} \left[ T(f_\theta(x_1), f_\theta(x_2)) \right] + \lambda \cdot \mathcal{R}(f_\theta(x_1), f_\theta(x_2)).
\end{equation}

Examples of JMI methods include contrastive learning methods such as SimCLR, which directly optimizes InfoNCE to maximize MI between views. More recent non-contrastive methods, such as Barlow Twins \citep{zbontar2021barlow} and VICReg \citep{bardes2021vicreg}, optimize an alignment term between augmented views together with a regularizer that approximates the marginal term from \cref{eq:dv_bound}. Whitening-based methods follow the same paradigm through a hard constraint rather than a penalty. W-MSE \citep{ermolov2021whitening} whitens the embeddings of each batch, enforcing an identity covariance, so the decorrelation that VICReg encourages through the regularizer in \cref{jmi_objective} is satisfied by construction and only the alignment term remains to be optimized.

\begin{figure}[t]
  \centering
  \includegraphics[width=1\textwidth]{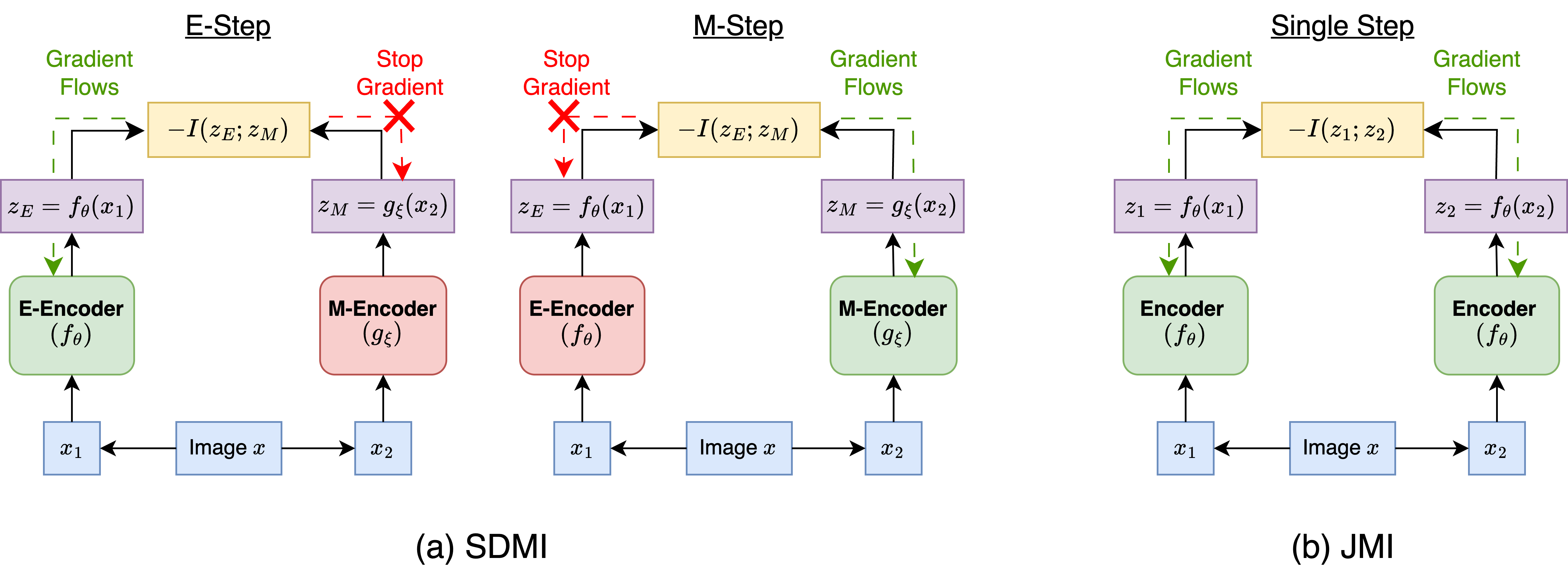}
  \caption{Canonical forms of our proposed paradigms: (a) SDMI alternates updates between two encoders using stop-gradients, while (b) JMI jointly updates both views with shared gradients.}
  \label{fig:sdmi_jmi}
\end{figure}

\subsection{From DV to Barlow Twins: A surrogate derivation}
\label{sec:section_3.4}

To show how non-contrastive methods can be seen as using feature-level regularization as in \cref{jmi_objective}, we demonstrate how the Barlow Twins loss could be derived from \cref{eq:dv_bound} using several straightforward approximations and assumptions, providing a direct connection between the Barlow Twins loss and mutual information maximization between views. To begin, we replace the DV bound's marginal term with its second-order Taylor approximation\footnote{The $\tfrac{1}{2}$ coefficient that appears in front of the variance term in the standard second-order Taylor expansion of the log-MGF is omitted here for simplicity, as it can be absorbed into a tuning hyperparameter.}:
\begin{equation}
\label{eq:taylor_dv_loss}
    \mathcal{L}_{\mathrm{Taylor\text{-}DV}} = -
    \underbrace{\mathbb{E}_{p(z_A,z_B)}[T(z_A,z_B)]}_{\text{Joint term}}
    +
    \underbrace{\mathbb{E}_{p(z_A)p(z_B)}[T(z_A,z_B)]}_{\text{Marginal mean term}}
    +
    \underbrace{\mathrm{Var}_{p(z_A)p(z_B)}[T(z_A,z_B)]}_{\text{Marginal variance term}}.
\end{equation}
Barlow Twins corresponds to the particular choice of the dot product scoring function,
\begin{equation}
\label{eq:barlow_twins_dot}
T(z^A, z^B) = \sum_{i=1}^d z^A_i z^B_i,
\end{equation}
which is an approximation to the optimal $T$ in \cref{eq:dv_bound}. Since batch normalization is normally applied, we also assume \(\mathbb{E}[z^A_i] \approx \mathbb{E}[z^B_i] \approx 0\), which effectively removes the marginal mean term from \cref{eq:taylor_dv_loss}, leaving the alignment and variance terms as the primary components.

Barlow Twins is usually expressed with the empirical cross-correlation matrix,
\[
C_{ij} = \frac{1}{N} \sum_{n=1}^N z^A_{n,i} \cdot z^B_{n,j},
\]
where alignment is encouraged via the diagonal \(C_{ii}\), and redundancy is penalized via the off-diagonals \(C_{ij}\), \(i \ne j\). To simplify the variance term in \cref{eq:taylor_dv_loss}, we expand it using the dot product in \cref{eq:barlow_twins_dot}. Under independent sampling, batch normalization pins the diagonal terms to a constant that does not depend on the encoder parameters, while the off-diagonal terms involve within-view second moments that track the cross-correlation matrix once the alignment term has drawn the two views together. Dropping the constant yields
\[
\mathrm{Var}\left[\sum_i z^A_i z^B_i\right] \approx \sum_{i \ne j} C_{ij}^2.
\]

Putting together these components yields a moment-based surrogate to the DV objective
\begin{equation}
    \label{eq:bt_taylor_dv_surrogate}
    \mathcal{L}_{\mathrm{Taylor\text{-}DV}} \approx -\sum_i C_{ii} + \sum_{i \ne j} C_{ij}^2,
\end{equation}
which closely matches the Barlow Twins loss. We provide the full derivation in \cref{app:moment_surrogates_for_DV_marginal}.

In summary, SDMI and JMI represent two principled optimization paradigms for maximizing MI. Our findings reveal that many architectural components in modern SSRL methods, previously introduced as heuristic choices, can instead be interpreted as structured consequences of optimizing MI. We illustrate the distinction between SDMI and JMI in \cref{fig:sdmi_jmi}, and provide a summary of representative SSRL methods in \cref{tab:method_classification}, together with their classification under the SDMI/JMI taxonomy, including whether they employ explicit MI objectives or surrogate regularizers.

\section{Experiments}
\label{sec:experiment}
This section empirically validates the theoretical structure of SDMI and JMI by instantiating their canonical forms and analyzing their behavior alongside representative SSRL methods. The purpose of this study is not to suggest that the canonical forms of SDMI and JMI are state-of-the-art SSRL methods. Instead, we use them as a simplified setting to understand the dynamics of MI training and representation quality, and to examine how the optimization principles derived from MI manifest in practice. We compare the canonical forms to more specialized and performant variants from the literature to shed light on the role of MI maximization in SSRL.

\subsection{Canonical SDMI and JMI prototypes}
\label{sec:canonical_forms}

To empirically validate the theoretical framework developed in \cref{sec:method}, we instantiate minimal, controlled implementations of the SDMI and JMI paradigms. These \textit{canonical forms} exclude auxiliary components such as momentum updates, predictor networks, or regularizers, and serve to isolate the optimization structure derived from the DV bound. As illustrated in \cref{fig:sdmi_jmi}, SDMI alternates updates between two encoders using stop-gradients, while JMI applies symmetric joint updates to both augmented views using a shared encoder. Both prototypes optimize the same MI objective defined below, enabling a direct comparison of their dynamics.

\textbf{Objective: Cosine-based DV bound}\quad While the DV bound is theoretically maximally tight when \(\mathcal{F}\) is a sufficiently broad class of functions, in practice, unrestricted neural critics $T$ often lead to high variance and unstable training behavior \citep{oord2018representation, poole2019variational, song2019understanding}. To ensure reliable estimation while preserving the validity of the DV bound, we restrict the critic function \(T\) to cosine similarity, \(T(z_A,z_B) = \frac{z_A \cdot z_B}{\|z_A\|_2 \|z_B\|_2}\), providing a stable, bounded, and scale-invariant surrogate. This choice is further motivated by its widespread use in SSRL objectives \citep{chen2020simple, he2020momentum, chen2020improved, chen2021empirical, grill2020bootstrap, chen2020exploring}, where it serves as a standard metric for comparing representations across augmented views. By restricting \(T\) to be the cosine similarity, we effectively optimize only over the representations of $Z_A$ and $Z_B$:
\begin{align}
    \label{eq:cos_dv_bound}
    I(Z_A; Z_B) \ge I_{\mathrm{DV}}(Z_A; Z_B)
    &\ge I_{\mathrm{cos\text{-}DV}}(Z_A; Z_B) \nonumber \\
    &= \mathbb{E}_{p(z_A, z_B)}\left[ T_{\mathrm{cos}}(z_A, z_B) \right]
       - \log \mathbb{E}_{p(z_A)p(z_B)}\left[ e^{T_{\mathrm{cos}}(z_A, z_B)} \right].
\end{align}
Although using $I_{\mathrm{cos\text{-}DV}}$ sacrifices some tightness, it provides a more stable estimator while remaining a lower bound of the MI objective.

\textbf{Practical approximation}\quad
To compute the marginal term in \cref{eq:cos_dv_bound} efficiently, we approximate the expectation using off-diagonal cross-pairs from a batch of size \(N\):
\[
\log\mathbb{E}_{p(z_A)p(z_B)}\left[e^{T_{\mathrm{cos}}(z_A, z_B)}\right] \approx \log\bigg(\frac{1}{N(N-1)} \sum_{\substack{i,j=1\\i\ne j}}^{N} e^{T_{\mathrm{cos}}(z_A^{(i)}, z_B^{(j)})}\bigg).
\]
Hence, our batchwise training objective takes the form:
\begin{equation}
    \label{eq:cos_dv_bound_loss}
    \mathcal{L}_{\mathrm{cos\text{-}DV}} = - \bigg[ \frac{1}{N} \sum_{i=1}^{N} T_{\mathrm{cos}}(z_A^{(i)}, z_B^{(i)}) - \log \bigg( \frac{1}{N(N-1)} \sum_{\substack{i,j=1\\i\ne j}}^{N} e^{T_{\mathrm{cos}}(z_A^{(i)}, z_B^{(j)})} \bigg) \bigg].
\end{equation}

Full pseudocode for the SDMI and JMI training procedures is provided in \cref{alg:sdmi_train_loop,alg:jmi_train_loop} of \cref{app:algorithms}.

\subsection{Experimental setup}
\label{sec:experimental_setup_toy}

\textbf{Datasets}\quad We use standard SSRL benchmarks, including CIFAR10/100 \citep{krizhevsky2009learning}, TinyImageNet, and ImageNet100 \citep{imagenet}. Additionally, for controlled experiments and visualization, we generate a toy dataset from a mixture of five isotropic Gaussian distributions centered at evenly spaced points on the unit circle. Each cluster center is defined by \(\mu_k = \Bigl(\cos\!\tfrac{2\pi k}{5},\,\sin\!\tfrac{2\pi k}{5}\Bigr), \quad k = 1,\dots,5,\) with samples drawn as \(x \sim \mathcal{N}(\mu_k, \sigma^2 I),\) where \(\sigma = 0.05\) and \(I\) is the \(2 \times 2\) identity matrix. Two augmented views are generated by perturbing \(x\) with independent Gaussian noise: \(x_1 = x + \epsilon_1,\quad x_2 = x + \epsilon_2,\quad \epsilon_1, \epsilon_2 \sim \mathcal{N}(0, \tau^2 I),\) where \(\tau = 0.1\). We generate \(N = 2500\) samples, with 500 samples per cluster.

\textbf{Implementation details}\quad
We implement the canonical SDMI prototype (\cref{fig:sdmi_jmi}(a)) with two independently initialized encoders trained via alternating E- and M-step updates. The JMI prototype (\cref{fig:sdmi_jmi}(b)) uses a single shared encoder updated jointly with symmetric gradients. All baseline methods use their standard architectures and objectives. Our canonical SDMI and JMI prototypes use ResNet-18~\citep{resnet} encoders for CIFAR10/100 and TinyImageNet, and ResNet-50 encoders for ImageNet100 \citep{imagenet}. For the Gaussian dataset, each method (including all baselines) uses a dedicated two-layer MLP encoder mapping \(\mathbb{R}^2 \to \mathbb{R}^3\): a linear layer (\(2 \to 64\)) with bias, batch normalization, and ReLU activation, followed by a second linear layer (\(64 \to 3\)) with batch normalization, with the output normalized to unit \(\ell_2\)-norm.

\begin{figure}[t]
  \centering
  \includegraphics[width=0.9\textwidth]{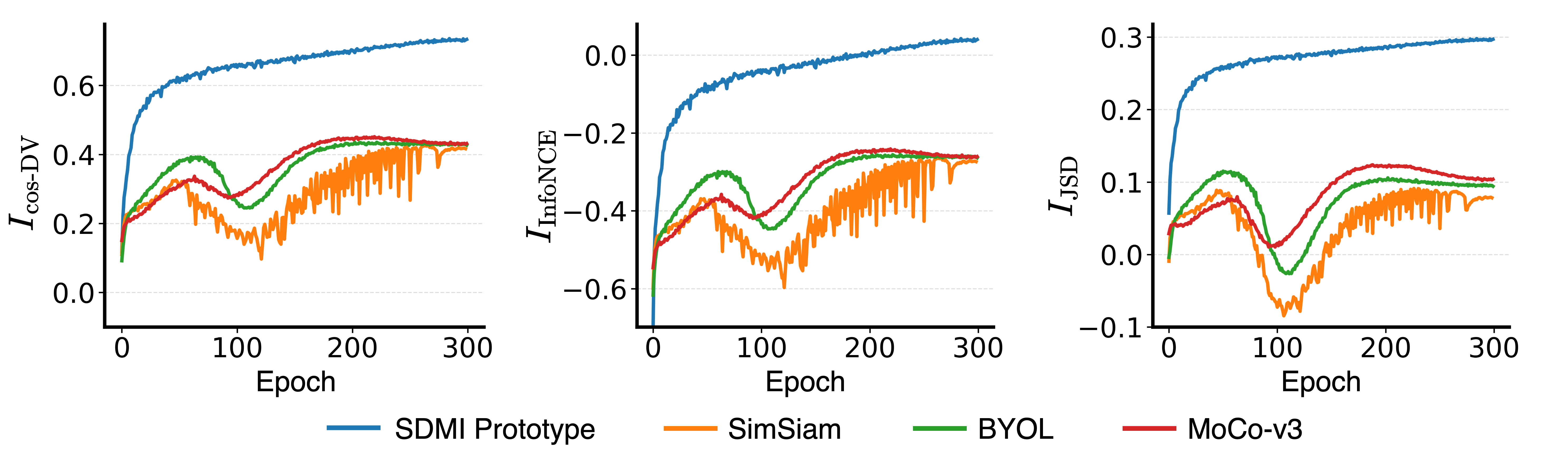}
  \includegraphics[width=0.9\textwidth]
  {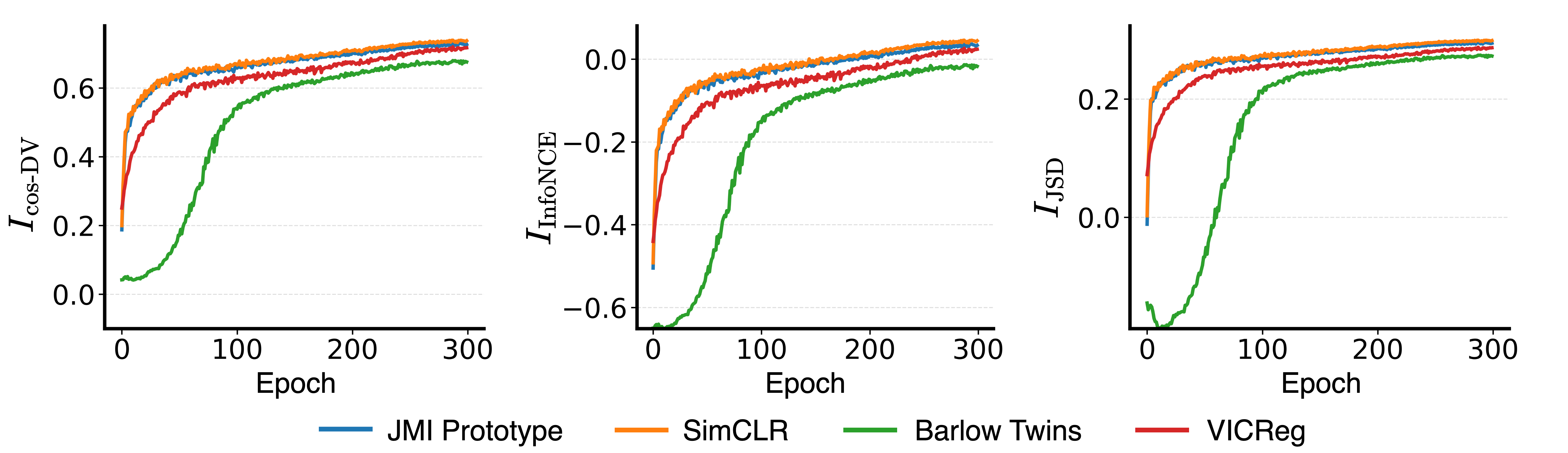}
  \caption{Estimated MI over CIFAR10 training for SDMI-based (top row) and JMI-based (bottom row) methods, using three estimators (cos-DV, InfoNCE, and JSD; left to right). Both paradigms exhibit consistent MI growth: SDMI curves feature early fluctuations before trending upward, while JMI estimates rise more uniformly and to much higher levels.}
  \label{fig:real_data_mi_estimates}
\end{figure}

\textbf{Mutual information estimation}\quad
To assess MI dynamics during training, we compute three variational bounds: the cos-DV bound (\(I_{\mathrm{cos\text{-}DV}}\)) from \cref{eq:cos_dv_bound}, the InfoNCE bound (\(I_{\mathrm{InfoNCE}}\)) \citep{oord2018representation, poole2019variational}, and the JSD bound (\(I_{\mathrm{JSD}}\)) \citep{hjelm2018learning}. Variational MI estimators are biased and can exhibit high variance in high dimensions \citep{poole2019variational, song2019understanding}, and with the critic restricted to cosine similarity each bound has a tightness that can differ across methods. We therefore read these estimates as qualitative trends, use the three structurally different bounds as cross-checks, and do not compare absolute values across estimators.

For JMI-based methods (JMI prototype, SimCLR, Barlow Twins, and VICReg), both augmented views are passed through the same encoder \(f_\theta\), and MI is computed between the representations:
\[
I^{(t)} = I\bigl(f_\theta^{(t)}(x_1),\,f_\theta^{(t)}(x_2)\bigr).
\]

For SDMI-based methods (SDMI prototype, SimSiam, BYOL, MoCo-v3), MI is measured between two asymmetric encoder branches. In the SDMI prototype, these are independently updated \(f_\theta\) and \(g_\xi\) encoders trained via alternating updates:
\[
I^{(t)} = I\bigl(f_\theta^{(t)}(x_1),\,g_\xi^{(t)}(x_2)\bigr).
\]
In BYOL and MoCo-v3, \(g_\xi\) is a momentum encoder updated via EMA. In SimSiam, which lacks a persistent target encoder, we instead treat the previous epoch's encoder state as the M-branch:
\[
I_\text{SimSiam}^{(t)} = I\bigl(f_\theta^{(t)}(x_1),\,f_\theta^{(t-1)}(x_2)\bigr),
\]
with \(I^{(0)} = -\infty\) by convention. 

The same estimation protocol and the same fixed critic are used for every method and dataset.

\subsection{Results}
\label{sec:results}

\textbf{Monotonic MI increase}\quad
\Cref{fig:real_data_mi_estimates} shows estimated MI over training for all methods across both paradigms on the CIFAR10 dataset. Since the SDMI prototype explicitly optimizes the cos-DV bound in \cref{eq:cos_dv_bound}, while MoCo-v3 optimizes InfoNCE, the JSD bound serves as an independent estimator not optimized by any method. Compared to the other SDMI methods, the SDMI prototype (top row) exhibits a near-perfect monotonic increase in MI throughout training. This is expected, as it explicitly optimizes the cos-DV bound (\cref{eq:cos_dv_bound}) using true EM-style alternating updates between two independently parameterized encoders. In contrast, methods like SimSiam, BYOL, and MoCo only approximate this behavior through their architectural heuristics, which lead to a noisy MI estimate and generally lower final MI levels. Nevertheless, all methods still exhibit an overall upward MI trend, confirming that they retain the underlying MI-maximization structure. Meanwhile, all JMI-based methods display smooth and consistently increasing MI curves, reflecting their symmetric joint-update optimization. We provide additional results on the Gaussian data in \cref{sec:synthetic_data}, confirming this trend in ideal conditions.

\begin{figure}[t]
  \centering
  \includegraphics[width=0.95\textwidth,  trim={8 8 8 2}, clip]
  {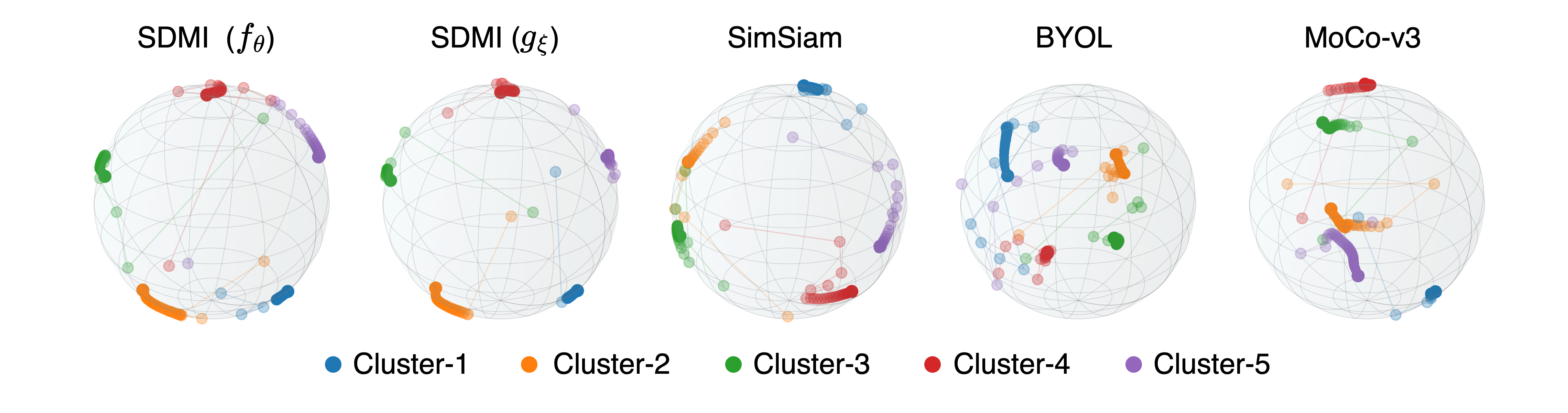}
  \includegraphics[width=0.95\textwidth,  trim={8 8 8 2}, clip]
  {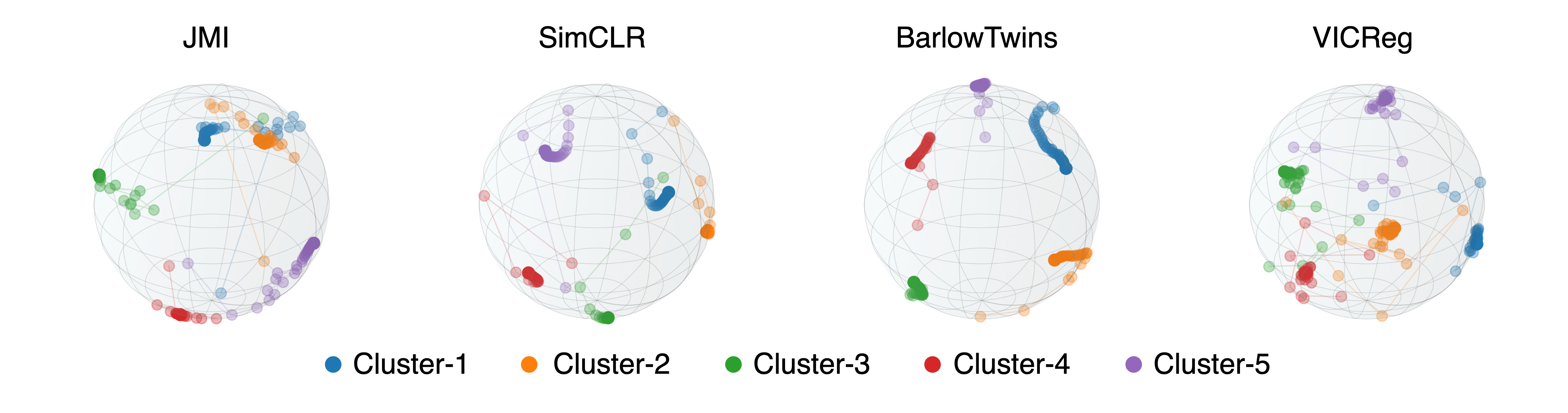}
  \caption{Embedding trajectories of the five Gaussian cluster centers. Opacity increases over training, showing how different methods progressively separate the clusters in embedding space.}
  \label{fig:toy_center_movement}
\end{figure}

\begin{table}[t]
    \centering
    \caption{Linear probing accuracy (\%) on four datasets. Mean ± std over 3 runs.}
    \begin{tabular}{@{}lcccc@{}}
        \toprule
        \textbf{Model} & \textbf{CIFAR10} & \textbf{CIFAR100} & \textbf{TinyImageNet} & \textbf{ImageNet100} \\
        \midrule
        \rowcolor{gray!10}SDMI prototype ($f_\theta$) & 88.61 {\scriptsize$\pm$0.13} & 57.37 {\scriptsize$\pm$0.38} & 33.30 {\scriptsize$\pm$0.58} & 70.73 {\scriptsize$\pm$0.57} \\
        \rowcolor{gray!10}SDMI prototype ($g_\xi$)    & 88.59 {\scriptsize$\pm$0.35} & 57.85 {\scriptsize$\pm$0.32} & 32.94 {\scriptsize$\pm$0.71} & 70.83 {\scriptsize$\pm$0.16} \\
        SimSiam                    & 89.72 {\scriptsize$\pm$0.18} & 60.45 {\scriptsize$\pm$0.60} & 19.19 {\scriptsize$\pm$0.69} & 78.23 {\scriptsize$\pm$0.58} \\
        BYOL                       & 91.28 {\scriptsize$\pm$0.16} & 63.11 {\scriptsize$\pm$0.21} & 32.77 {\scriptsize$\pm$0.10} & 81.09 {\scriptsize$\pm$0.61} \\
        MoCo-v3                    & 91.10 {\scriptsize$\pm$0.16} & 58.90 {\scriptsize$\pm$0.32} & 32.18 {\scriptsize$\pm$0.55} & 76.86 {\scriptsize$\pm$0.74} \\
        \midrule
        \rowcolor{gray!10}JMI Prototype & 88.01 {\scriptsize$\pm$0.48} & 57.22 {\scriptsize$\pm$0.56} & 32.23 {\scriptsize$\pm$0.52} & 73.41 {\scriptsize$\pm$0.36} \\
        SimCLR                     & 87.24 {\scriptsize$\pm$0.37} & 55.32 {\scriptsize$\pm$0.46} & 33.79 {\scriptsize$\pm$0.31} & 75.31 {\scriptsize$\pm$0.76} \\
        Barlow Twins               & 85.56 {\scriptsize$\pm$0.71} & 51.91 {\scriptsize$\pm$0.49} & 30.26 {\scriptsize$\pm$0.12} & 78.96 {\scriptsize$\pm$0.30} \\
        VICReg                     & 85.49 {\scriptsize$\pm$1.03} & 54.00 {\scriptsize$\pm$0.34} & 32.03 {\scriptsize$\pm$0.32} & 78.86 {\scriptsize$\pm$0.23} \\
        \bottomrule
    \end{tabular}
    \label{tab:linear_probe_results}
\end{table}

\textbf{Cluster center trajectories in embedding space}\quad To visualize how well the representation space separates underlying structure, we use the Gaussian dataset and track the movement of all five cluster centers during training (\cref{fig:toy_center_movement}). We quantify separation via the nearest-neighbor (NN) angle gap, the mean angular distance to the closest other center. The JMI prototype achieves the largest separation, reaching an average NN gap of $\approx 88^\circ$, with the closest JMI-based competitor at $\approx 82^\circ$. The SDMI prototype also shows strong separation, with an average NN angle gap of \(\approx74^\circ\), compared to \(\approx50^\circ\) for the strongest SDMI-based competitor. Detailed metrics and comparative analysis are presented in \cref{sec:toy_experiment}.

\textbf{Linear probing}\quad To assess the quality of learned representations for downstream tasks, we perform linear probing on real-world datasets. We train encoders, freeze them, and then train a linear classifier head using cross-entropy~\citep{tian2020contrastive}. As shown in \cref{tab:linear_probe_results}, the prototypes are competitive within their respective paradigms, with the JMI prototype strongest among JMI methods on CIFAR10 and CIFAR100 and the SDMI prototype strongest among SDMI methods on TinyImageNet. Larger gaps remain on ImageNet100, approximately 10\% for SDMI and 6\% for JMI relative to the strongest baselines, where the SDMI prototype also used a reduced batch size (\cref{app:imagenet100_tinyimagenet}). No single method outperforms all others consistently. See \cref{app:implementation_details} for implementation details.

It is worth noting that our canonical SDMI and JMI models are intentionally minimal, showing that theory-driven models can provide strong baselines without the need for empirically driven architectural tweaks like predictor heads, EMA, or regularization. Existing SSRL methods can be viewed as augmenting such minimal forms with task-specific architectural improvements. Our work focuses on explanation rather than optimization.

\phantomsection\label{sec:predictor_ablation}
\textbf{Predictors as DV-bound marginal surrogates}\quad

\begin{figure}[t]
  \centering
  \includegraphics[width=1\textwidth]{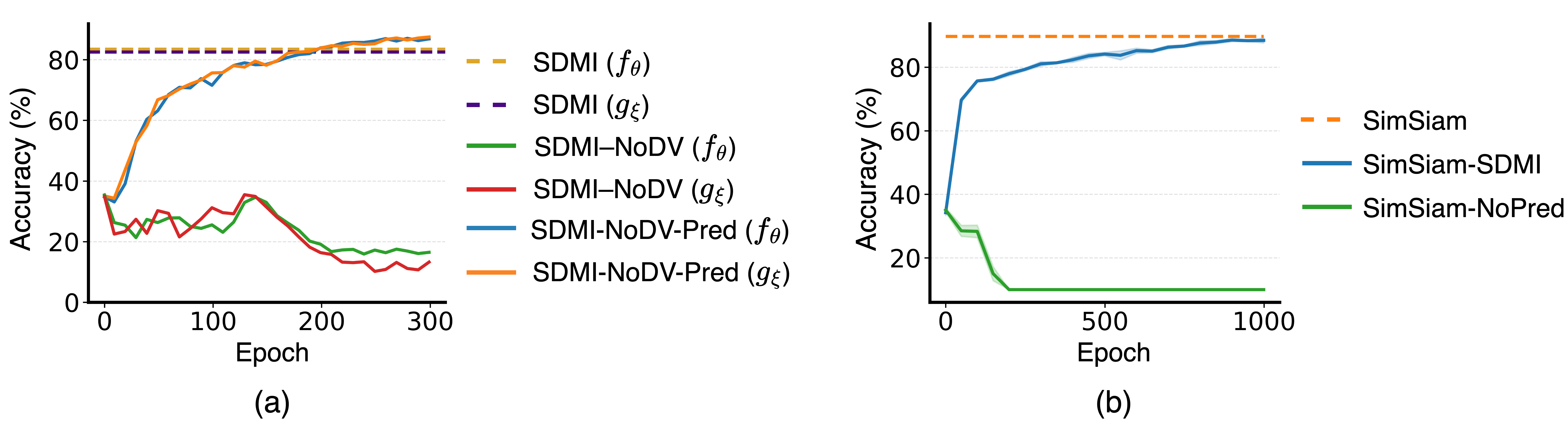}
  \caption{\textbf{(a) Linear probing accuracy of SDMI prototype under ablations on CIFAR10.} Removing the marginal term from the DV bound leads to representational collapse in both the \(f_\theta\) and \(g_\xi\) encoders, denoted as SDMI-NoDV (\(f_\theta\)) and SDMI-NoDV (\(g_\xi\)), respectively, despite the use of alternating (EM-style) optimization, confirming the necessity of the marginal regularization term. Remarkably, adding trainable predictors during the E-step and M-step, SDMI-NoDV-Pred (\(f_\theta\)) and SDMI-NoDV-Pred (\(g_\xi\)), while still omitting the marginal term, entirely prevents collapse and recovers strong performance. Dashed lines represent baseline accuracy of the pure SDMI prototype with all components intact.\quad \textbf{(b) SimSiam variants under different loss functions and predictor configurations on CIFAR10.} Removing the predictor leads to collapse, confirming that SimSiam's cosine loss alone lacks marginal regularization. Replacing the loss with the explicit cos-DV objective (SimSiam-SDMI) restores performance without requiring a predictor.}
  \label{fig:predictor_ablation}
\end{figure}

Predictor networks and stop-gradients are widely recognized as essential components in SDMI-based SSRL methods \citep{chen2020exploring, balestriero2023cookbook, jha2024common}, with several works analyzing their role in preventing collapse \citep{zhang2022does, tian2021understanding, srinath2023implicit, shi2020run, wang2021towards}. In \cref{sec:EM_style_MI_maximization}, we show that stop-gradient enables the EM-style alternating optimization. \citet{zhang2022does} prove that the predictor prevents collapse by decomposing its gradient into center and residual components, showing it induces de-centering and dimensional de-correlation, mechanisms equivalent to those from negative samples in contrastive learning. We complement this understanding with a controlled ablation study on CIFAR10 using linear probing (\cref{fig:predictor_ablation}(a)), systematically adding and removing components from the SDMI prototype. Removing the marginal term from \cref{eq:dv_bound} results in representational collapse, despite EM-style updates, confirming the necessity of a complete MI objective. More generally, in the absence of the marginal regularization term, encoders converge to trivial solutions.

Remarkably, adding a trainable predictor to the \(f_\theta\) encoder during the E-step and to the \(g_\xi\) encoder during the M-step, while still omitting the marginal term, entirely prevents collapse. This behavior suggests that predictors function as implicit surrogates for the log-partition (marginal) term, validating our analysis in \cref{sec:byol_and_simsiam} that predictors recover the missing normalization component in the MI maximization objective.

Using SimSiam as an example, we now examine what happens when a non-contrastive method within the SDMI paradigm is given an explicit MI objective. First, we remove the predictor from SimSiam while retaining its alternating update scheme. This variant collapses, since minimizing cosine similarity is equivalent to optimizing only the joint term of the DV bound (\cref{eq:dv_bound}), with no marginal correction to prevent trivial solutions. Next, we replace SimSiam's heuristic loss with the cos-DV bound-based loss from \cref{eq:cos_dv_bound_loss}, while still using a single-encoder alternating schedule but without the predictor. We call this variant SimSiam-SDMI. Remarkably, SimSiam-SDMI not only avoids collapse, but also recovers nearly the original linear probe accuracy (baseline: \(89.68 \pm 0.35\%\), SimSiam-SDMI: \(88.43 \pm 0.78\%\)). \Cref{fig:predictor_ablation}(b) summarizes these trends across all three variants.

\textbf{Discussion}\quad
Interestingly, while the SDMI prototype achieves the highest MI under all three bounds and the most separated clusters among SDMI methods, this does not translate directly into higher downstream performance. MI is invariant to invertible transformations of the representation space, so it measures how much information is preserved but not the geometry in which that information is arranged, whereas linear probing rewards linearly separable structure \citep{tschannen2020mutual}. Furthermore, representations with similar MI between views can retain different subsets of the input information (\cref{app:input_informativeness}), and components such as predictors and statistical regularizers shape representation geometry in ways that benefit linear evaluation but are not reflected in MI. This suggests that maximizing MI, though necessary to prevent representational collapse, is not by itself sufficient for optimal SSRL performance.

MI should thus be viewed as a foundation rather than the ultimate objective of SSRL. Our results show that the optimization paradigm, SDMI or JMI, and the strategies and components it uses, determine how the MI objective is approximated and, in turn, the usefulness of the learned features for downstream tasks. In summary, \textit{how} MI is optimized matters as much as \textit{how much} MI is achieved. By formalizing the SDMI and JMI paradigms and identifying their essential components, our taxonomy provides a roadmap for future research. We recommend that future efforts prioritize designing better strategies and architectural components tailored to each optimization paradigm, thereby better bridging the gap between MI maximization and downstream task performance.

\subsection{MI estimation}
\label{sec:synthetic_data}

\begin{figure}[t]
  \centering
  \includegraphics[width=\textwidth, trim = {0 0 0 20}, clip]{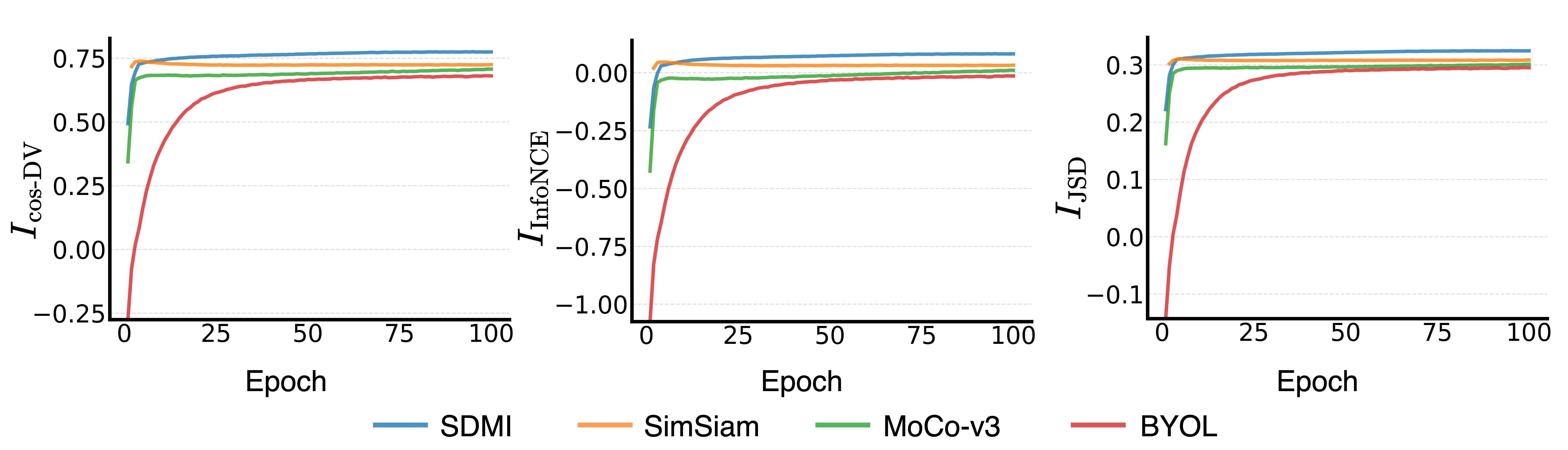} \\\includegraphics[width=\textwidth, trim = {0 0 0 20}, clip]
  {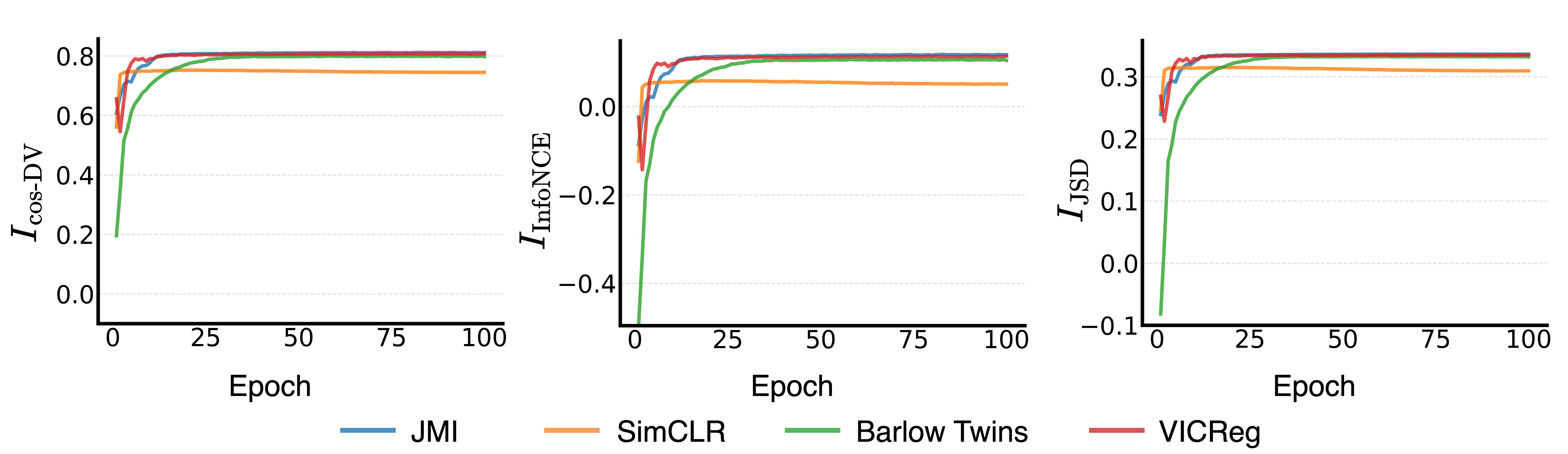}
  \caption{Estimated MI over training using cos-DV, InfoNCE, and JSD bounds for both SDMI methods (top) and JMI methods (bottom). All three estimators show approximately monotonic growth for all methods under both paradigms.}\label{fig:mi_estimates_sdmi_jmi}
\end{figure}

To better understand the dynamics of MI maximization without confounding factors from complex image data, we repeat our MI-tracking experiment on the controlled synthetic dataset and toy MLP encoder described in \cref{sec:experimental_setup_toy}. In this setting, we can isolate the effect of optimization since the ground-truth data distribution is simple and noise is well-characterized. We compute the same three variational bounds in a deterministic setting (i.e., a single batch update): the cos-DV bound (\(I_{\mathrm{cos\text{-}DV}}\)) from \cref{eq:cos_dv_bound}, the InfoNCE bound (\(I_{\mathrm{InfoNCE}}\)) \citep{oord2018representation, poole2019variational}, and the JSD bound (\(I_{\mathrm{JSD}}\)) \citep{hjelm2018learning}. At each epoch, we compute all three MI estimates on frozen encoder outputs from a validation set consisting of 2,500 data points. As shown in \cref{fig:mi_estimates_sdmi_jmi}, all three bounds across all methods show similar, near-monotonic MI increase during training.

\section{Conclusion}
\label{sec:conclusion}
In this work, we revisited SSRL from first principles, grounding our analysis in a variational MI optimization lens. By deriving two distinct training paradigms, SDMI and JMI, we showed that many design choices in SSRL architectures are not merely empirical conveniences, but arise as principled consequences of, or surrogates for, optimizing an MI lower bound. By unifying a broad class of existing SSRL methods under the theoretical lens, our analysis offers an alternative understanding of the mechanisms that drive successful representation learning and guides the principled design of future SSRL algorithms.

\subsubsection*{Limitations}

While our framework offers a principled view of SSRL via MI maximization, our experiments are limited to image datasets. Extending the analysis to other data modalities such as text, audio, or multimodal settings would strengthen the generalizability of our theoretical insights and is a promising direction for future work, as is scaling the empirical comparison to ImageNet-1k and extending evaluation beyond linear probing to fine-tuning and to dense prediction tasks such as detection and segmentation.

\subsubsection*{Broader Impact Statement}

By clarifying the principles behind self-supervised learning, this work may support more robust and interpretable model design. Though theoretical, our findings could influence the development of trustworthy AI systems in socially impactful domains.

\subsubsection*{Reproducibility Statement}
\label{sec:reproducibility}
We have made significant efforts to ensure that our results are fully reproducible. \Cref{sec:method} formally derives the proposed SDMI and JMI paradigms and lists all assumptions, with complete proofs in \cref{app:further_analysis} and training procedures in \cref{app:algorithms}. \Cref{sec:experimental_setup_toy} describes our experimental setup, including synthetic data generation and evaluation protocols, while \cref{app:implementation_details} provides full implementation details, compute resources, random seed settings, and hyperparameter configurations for CIFAR10/100, TinyImageNet, and ImageNet100. Hyperparameter sweeps and selected settings are reported in \cref{tab:hyperparams}–\cref{tab:imagenet100_tinyimagenet_hyperparams}, and additional results and ablations are presented in \cref{app:additional_experimental_results}. The implementation containing all code for model training, evaluation, and MI estimation is publicly available at \href{https://github.com/AkhlaqurRahmanSabby/SSRL-as-MI-Maximization}{\texttt{github.com/AkhlaqurRahmanSabby/SSRL-as-MI-Maximization}} to enable exact replication.
\clearpage

\bibliographystyle{tmlr}
\bibliography{bib}

@inproceedings{grill2020bootstrap,
 author = {Grill, Jean-Bastien and Strub, Florian and Altch\'{e}, Florent and Tallec, Corentin and Richemond, Pierre and Buchatskaya, Elena and Doersch, Carl and Avila Pires, Bernardo and Guo, Zhaohan and Gheshlaghi Azar, Mohammad and Piot, Bilal and Kavukcuoglu, Koray and Munos, Remi and Valko, Michal},
 booktitle = {Advances in Neural Information Processing Systems},
 pages = {21271--21284},
 title = {{Bootstrap Your Own Latent - A New Approach to Self-Supervised Learning}},
 volume = {33},
 year = {2020}
}

@INPROCEEDINGS{imagenet,
  author={Deng, Jia and Dong, Wei and Socher, Richard and Li, Li-Jia and Kai Li and Li Fei-Fei},
  booktitle={2009 IEEE Conference on Computer Vision and Pattern Recognition}, 
  title={{ImageNet: A large-scale hierarchical image database}}, 
  year={2009},
  pages={248-255},
  doi={10.1109/CVPR.2009.5206848}}

@InProceedings{chen2020exploring,
    author    = {Chen, Xinlei and He, Kaiming},
    title     = {Exploring Simple Siamese Representation Learning},
    booktitle = {Proceedings of the IEEE/CVF Conference on Computer Vision and Pattern Recognition},
    year      = {2021},
    pages     = {15750-15758}
}

@InProceedings{chen2020simple,
  title = 	 {A Simple Framework for Contrastive Learning of Visual Representations},
  author =       {Chen, Ting and Kornblith, Simon and Norouzi, Mohammad and Hinton, Geoffrey},
  booktitle = 	 {Proceedings of the 37th International Conference on Machine Learning},
  pages = 	 {1597--1607},
  year = 	 {2020},
  volume = 	 {119},
}

@InProceedings{poole2019variational,
  title = 	 {On Variational Bounds of Mutual Information},
  author =       {Poole, Ben and Ozair, Sherjil and Van Den Oord, Aaron and Alemi, Alex and Tucker, George},
  booktitle = 	 {Proceedings of the 36th International Conference on Machine Learning},
  pages = 	 {5171--5180},
  year = 	 {2019},
  volume = 	 {97},
}

@article{oord2018representation,
  title={Representation learning with contrastive predictive coding},
  author={Oord, Aaron van den and Li, Yazhe and Vinyals, Oriol},
  journal={arXiv:1807.03748},
  year={2018}
}

@inproceedings{
sui2023self,
title={Self-supervised Representation Learning from Random Data Projectors},
author={Yi Sui and Tongzi Wu and Jesse C. Cresswell and Ga Wu and George Stein and Xiao Shi Huang and Xiaochen Zhang and Maksims Volkovs},
booktitle={The Twelfth International Conference on Learning Representations},
year={2024},
}

@inproceedings{
zhang2022does,
title={{How Does SimSiam Avoid Collapse Without Negative Samples? A Unified Understanding with Self-supervised Contrastive Learning}},
author={Chaoning Zhang and Kang Zhang and Chenshuang Zhang and Trung X. Pham and Chang D. Yoo and In So Kweon},
booktitle={International Conference on Learning Representations},
year={2022},
}

@article{balestriero2023cookbook,
  title={A cookbook of self-supervised learning},
      author={Randall Balestriero and Mark Ibrahim and Vlad Sobal and Ari Morcos and Shashank Shekhar and Tom Goldstein and Florian Bordes and Adrien Bardes and Gregoire Mialon and Yuandong Tian and Avi Schwarzschild and Andrew Gordon Wilson and Jonas Geiping and Quentin Garrido and Pierre Fernandez and Amir Bar and Hamed Pirsiavash and Yann LeCun and Micah Goldblum},
  journal={arXiv:2304.12210},
  year={2023}
}

@article{shi2020run,
  title={{Run away from your teacher: Understanding BYOL by a novel self-supervised approach}},
  author={Shi, Haizhou and Luo, Dongliang and Tang, Siliang and Wang, Jian and Zhuang, Yueting},
  journal={arXiv:2011.10944},
  year={2020}
}

@InProceedings{chen2021empirical,
    author    = {Chen, Xinlei and Xie, Saining and He, Kaiming},
    title     = {An Empirical Study of Training Self-Supervised Vision Transformers},
    booktitle = {Proceedings of the IEEE/CVF International Conference on Computer Vision},
    year      = {2021},
    pages     = {9640-9649}
}

@InProceedings{he2020momentum,
author = {He, Kaiming and Fan, Haoqi and Wu, Yuxin and Xie, Saining and Girshick, Ross},
title = {Momentum Contrast for Unsupervised Visual Representation Learning},
booktitle = {Proceedings of the IEEE/CVF Conference on Computer Vision and Pattern Recognition (CVPR)},
year = {2020}
}

@InProceedings{zbontar2021barlow,
  title = 	 {Barlow Twins: Self-Supervised Learning via Redundancy Reduction},
  author =       {Zbontar, Jure and Jing, Li and Misra, Ishan and LeCun, Yann and Deny, Stephane},
  booktitle = 	 {Proceedings of the 38th International Conference on Machine Learning},
  pages = 	 {12310--12320},
  year = 	 {2021},
  volume = 	 {139},
}

@InProceedings{caron2021emerging,
    author    = {Caron, Mathilde and Touvron, Hugo and Misra, Ishan and J\'egou, Herv\'e and Mairal, Julien and Bojanowski, Piotr and Joulin, Armand},
    title     = {Emerging Properties in Self-Supervised Vision Transformers},
    booktitle = {Proceedings of the IEEE/CVF International Conference on Computer Vision},
    year      = {2021},
    pages     = {9650-9660}
}

@article{chen2020improved,
  title={Improved baselines with momentum contrastive learning},
  author={Chen, Xinlei and Fan, Haoqi and Girshick, Ross and He, Kaiming},
  journal={arXiv:2003.04297},
  year={2020}
}

@InProceedings{resnet,
author = {He, Kaiming and Zhang, Xiangyu and Ren, Shaoqing and Sun, Jian},
title = {Deep Residual Learning for Image Recognition},
booktitle = {Proceedings of the IEEE Conference on Computer Vision and Pattern Recognition},
month = {June},
year = {2016}
}

@InProceedings{belghazi2018mine,
  title = 	 {Mutual Information Neural Estimation},
  author =       {Belghazi, Mohamed Ishmael and Baratin, Aristide and Rajeshwar, Sai and Ozair, Sherjil and Bengio, Yoshua and Courville, Aaron and Hjelm, Devon},
  booktitle = 	 {Proceedings of the 35th International Conference on Machine Learning},
  pages = 	 {531--540},
  year = 	 {2018},
  volume = 	 {80},
}

@inproceedings{
hjelm2018learning,
title={Learning deep representations by mutual information estimation and maximization},
author={R Devon Hjelm and Alex Fedorov and Samuel Lavoie-Marchildon and Karan Grewal and Phil Bachman and Adam Trischler and Yoshua Bengio},
booktitle={International Conference on Learning Representations},
year={2019},
}

@inproceedings{
bardes2021vicreg,
title={{VICReg: Variance-Invariance-Covariance Regularization for Self-Supervised Learning}},
author={Adrien Bardes and Jean Ponce and Yann LeCun},
booktitle={International Conference on Learning Representations},
year={2022},
}

@InProceedings{tian2021understanding,
  title = 	 {Understanding self-supervised learning dynamics without contrastive pairs},
  author =       {Tian, Yuandong and Chen, Xinlei and Ganguli, Surya},
  booktitle = 	 {Proceedings of the 38th International Conference on Machine Learning},
  pages = 	 {10268--10278},
  year = 	 {2021},
  volume = 	 {139},
}

@InProceedings{ermolov2021whitening,
  title = 	 {Whitening for Self-Supervised Representation Learning},
  author =       {Ermolov, Aleksandr and Siarohin, Aliaksandr and Sangineto, Enver and Sebe, Nicu},
  booktitle = 	 {Proceedings of the 38th International Conference on Machine Learning},
  pages = 	 {3015--3024},
  year = 	 {2021},
  volume = 	 {139},
}

@inproceedings{srinath2023implicit,
 author = {Srinath Halvagal, Manu and Laborieux, Axel and Zenke, Friedemann},
 booktitle = {Advances in Neural Information Processing Systems},
 pages = {63409--63436},
 title = {{Implicit variance regularization in non-contrastive SSL}},
 volume = {36},
 year = {2023}
}

@article{jha2024common,
  title={The Common Stability Mechanism behind most Self-Supervised Learning Approaches},
  author={Jha, Abhishek and Blaschko, Matthew B and Asano, Yuki M and Tuytelaars, Tinne},
  journal={arXiv:2402.14957},
  year={2024}
}

@inproceedings{caron2020unsupervised,
 author = {Caron, Mathilde and Misra, Ishan and Mairal, Julien and Goyal, Priya and Bojanowski, Piotr and Joulin, Armand},
 booktitle = {Advances in Neural Information Processing Systems},
 pages = {9912--9924},
 title = {Unsupervised Learning of Visual Features by Contrasting Cluster Assignments},
 volume = {33},
 year = {2020}
}

@inproceedings{barber2004algorithm,
 author = {Barber, David and Agakov, Felix},
 booktitle = {Advances in Neural Information Processing Systems},
 pages = {},
 publisher = {MIT Press},
 title = {{Information Maximization in Noisy Channels : A Variational Approach}},
 volume = {16},
 year = {2003}
}

@article{nguyen2010estimating,
  title={Estimating divergence functionals and the likelihood ratio by convex risk minimization},
  author={Nguyen, XuanLong and Wainwright, Martin J and Jordan, Michael I},
  journal={IEEE Transactions on Information Theory},
  volume={56},
  number={11},
  pages={5847--5861},
  year={2010},
  publisher={IEEE}
}

@inproceedings{
song2019understanding,
title={Understanding the Limitations of Variational Mutual Information Estimators},
author={Jiaming Song and Stefano Ermon},
booktitle={International Conference on Learning Representations},
year={2020},
}

@INPROCEEDINGS{tishby2015deep,
  author={Tishby, Naftali and Zaslavsky, Noga},
  booktitle={2015 IEEE Information Theory Workshop}, 
  title={Deep learning and the information bottleneck principle}, 
  year={2015},
  pages={1-5},
  doi={10.1109/ITW.2015.7133169}}

@article{krizhevsky2009learning,
  title={Learning multiple layers of features from tiny images},
  author={Krizhevsky, Alex and Hinton, Geoffrey},
  year={2009},
  publisher={University of Toronto}
}

@inproceedings{tishby2000information,
  author = {Tishby, Naftali and Pereira, Fernando C. and Bialek, William},
  booktitle = {Proc. of the 37-th Annual Allerton Conference on Communication, Control and Computing},
  pages = {368-377},
  title = {The information bottleneck method},
  year = 1999
}

@article{tsai2021note,
  title={{A note on connecting Barlow twins with negative-sample-free contrastive learning}},
  author={Tsai, Yao-Hung Hubert and Bai, Shaojie and Morency, Louis-Philippe and Salakhutdinov, Ruslan},
  journal={arXiv:2104.13712},
  year={2021}
}

@InProceedings{gretton2005measuring,
author="Gretton, Arthur
and Bousquet, Olivier
and Smola, Alex
and Sch{\"o}lkopf, Bernhard",
title="{Measuring Statistical Dependence with Hilbert-Schmidt Norms}",
booktitle="Algorithmic Learning Theory",
year="2005",
publisher="Springer Berlin Heidelberg",
pages="63--77",
isbn="978-3-540-31696-1"
}

@inproceedings{shwartz2023information,
 author = {Shwartz-Ziv, Ravid and Balestriero, Randall and Kawaguchi, Kenji and Rudner, Tim G. J. and LeCun, Yann},
 booktitle = {Advances in Neural Information Processing Systems},
 pages = {33965--33998},
 title = {An Information Theory Perspective on Variance-Invariance-Covariance Regularization},
 volume = {36},
 year = {2023}
}

@article{wang2021towards,
  title={Towards demystifying representation learning with non-contrastive self-supervision},
  author={Wang, Xiang and Chen, Xinlei and Du, Simon S and Tian, Yuandong},
  journal={arXiv:2110.04947},
  year={2021}
}

@inproceedings{tian2020contrastive,
  title={Contrastive multiview coding},
  author={Tian, Yonglong and Krishnan, Dilip and Isola, Phillip},
  booktitle={Computer Vision--ECCV 2020: 16th European Conference, Glasgow, UK, August 23--28, 2020, Proceedings, Part XI 16},
  pages={776--794},
  year={2020},
  organization={Springer}
}

@inproceedings{
thakoor2021large,
title={Large-Scale Representation Learning on Graphs via Bootstrapping},
author={Shantanu Thakoor and Corentin Tallec and Mohammad Gheshlaghi Azar and Mehdi Azabou and Eva L Dyer and Remi Munos and Petar Veli{\v{c}}kovi{\'c} and Michal Valko},
booktitle={International Conference on Learning Representations},
year={2022},
}

@inproceedings{liu2022self,
 author = {Liu, Xin and Wang, Zhongdao and Li, Ya-Li and Wang, Shengjin},
 booktitle = {Advances in Neural Information Processing Systems},
 pages = {34091--34105},
 title = {Self-Supervised Learning via Maximum Entropy Coding},
 volume = {35},
 year = {2022}
}

@inproceedings{
tan2023information,
title={Information Flow in Self-Supervised Learning},
author={Zhiquan Tan and Jingqin Yang and Weiran Huang and Yang Yuan and Yifan Zhang},
booktitle={Forty-first International Conference on Machine Learning},
year={2024},
}

@article{lecun2015deep,
  title={Deep learning},
  author={LeCun, Yann and Bengio, Yoshua and Hinton, Geoffrey},
  journal={Nature},
  volume={521},
  number={7553},
  pages={436--444},
  year={2015},
  publisher={Nature Publishing Group UK London}
}

@article{bengio2013representation,
  title={Representation learning: A review and new perspectives},
  author={Bengio, Yoshua and Courville, Aaron and Vincent, Pascal},
  journal={IEEE Transactions on Pattern Analysis and Machine Intelligence},
  volume={35},
  number={8},
  pages={1798--1828},
  year={2013},
  publisher={IEEE}
}

@inproceedings{tschannen2020mutual,
  title={On Mutual Information Maximization for Representation Learning},
  author={Tschannen, Michael and Djolonga, Josip and Rubenstein, Paul K and Gelly, Sylvain and Lucic, Mario},
  booktitle={International Conference on Learning Representations},
  year={2020}
}

@inproceedings{nguyen2023lvm,
  title={{LVM-Med}: Learning Large-Scale Self-Supervised Vision Models for Medical Imaging via Second-order Graph Matching},
  author={Nguyen, Duy M. H. and Nguyen, Hoang and Diep, Nghiem T. and Pham, Tan N. and Cao, Tri and Nguyen, Binh T. and Swoboda, Paul and Ho, Nhat and Albarqouni, Shadi and Xie, Pengtao and Sonntag, Daniel and Niepert, Mathias},
  booktitle={Advances in Neural Information Processing Systems},
  year={2023}
}

@inproceedings{wen2022mechanism,
  title={The Mechanism of Prediction Head in Non-contrastive Self-supervised Learning},
  author={Wen, Zixin and Li, Yuanzhi},
  booktitle={Advances in Neural Information Processing Systems},
  year={2022}
}

\appendix
\crefalias{section}{appendix}
\crefalias{subsection}{appendix}
\crefalias{subsubsection}{appendix}
\clearpage
\section{Further analysis}
\label{app:further_analysis}

\subsection{Block-coordinate ascent in MI bounds}
\label{app:block_coordinate}

We provide the formal proof for the proposition stated in \cref{sec:section_3.1}, establishing the theoretical foundation for monotonic MI increase under alternating optimization in the SDMI paradigm.

\paragraph{Proof.}
Fix \(\xi^{(k)}\). Since \(J(\cdot;\xi^{(k)})\) is concave, a gradient ascent step on \(\theta\) with sufficiently small step size guarantees
\begin{equation}
    J(\theta^{(k+1)};\xi^{(k)}) \ge J(\theta^{(k)};\xi^{(k)}).
\end{equation}
By smoothness of \(M\) and the bound \(\|\nabla_\theta M\| \le \varepsilon\), we have
\begin{equation}
    \left| M(\theta^{(k+1)};\xi^{(k)}) - M(\theta^{(k)};\xi^{(k)}) \right| = \mathcal{O}(\varepsilon),
\end{equation}
which yields
\begin{equation}
    \mathcal{L}(\theta^{(k+1)}, \xi^{(k)}) \ge \mathcal{L}(\theta^{(k)}, \xi^{(k)}) - \mathcal{O}(\varepsilon).
\end{equation}
An identical argument applies to the \(\xi\)-update. Chaining the two completes the proof. \qed

\paragraph{Remark.} The proof makes explicit that the result is a perturbation bound. The concavity assumption yields ascent in the joint term for sufficiently small step sizes, while the $\varepsilon$-bound on the marginal gradient limits how much of that ascent can be offset; strict monotonicity is recovered only at $\varepsilon = 0$. See \cref{sec:section_3.1} for a discussion of when these assumptions can be expected to hold in practice.

\subsection{Analyzing other variational bounds}
\label{app:other_variational_bounds}

Extending the analysis from \cref{sec:section_3.1}, we examine other commonly used variational MI bounds in SSRL, namely the InfoNCE and JSD bounds mentioned in \cref{sec:section_2.2}, demonstrating that our framework generalizes beyond the DV bound.

\subsubsection{InfoNCE}

Recall that the InfoNCE loss between two representations \(Z_A\) and \(Z_B\) takes the form
\begin{align}
    \label{eq:infonce_loss}
    \mathcal{L}_{\mathrm{InfoNCE}}
    &= - \mathbb{E}_{p(z_A,z_B)}\!\left[\log\!\left( \frac{e^{T(z_A,z_B)}}{\sum_{z_B'} e^{T(z_A,z_B')}} \right)\right] \nonumber \\
    &= - \mathbb{E}_{p(z_A,z_B)}\!\left[ T(z_A,z_B) - \log\!\left( \sum_{z_B'} e^{T(z_A,z_B')} \right) \right],
\end{align}
where \(T\) is a similarity function and the inner sum runs over the negative samples \(z_B'\) in a batch.

This loss can be interpreted as a lower bound on MI between \(Z_A\) and \(Z_B\) \citep{poole2019variational}, such that
\begin{equation}
    \label{eq:infonce_bound}
    I_{\mathrm{InfoNCE}}(Z_A; Z_B) = \mathbb{E}_{p(z_A, z_B)}\!\left[ T(z_A, z_B) \right]
    - \mathbb{E}_{p(z_A)} \!\left[ \log \mathbb{E}_{p(z_B)} \!\left[e^{T(z_A, z_B)} \right] \right]
    + \log N,
\end{equation}
where \(N\) is the number of negative samples.

Both the DV and the InfoNCE bounds follow the general structure of a joint term minus a marginal term. The only structural difference is in how the logarithm is aggregated: the DV bound aggregates globally before applying the logarithm,
\[
    \log \mathbb{E}_{p(z_A)p(z_B)}\!\left[e^{T(z_A,z_B)}\right] \quad \text{(global aggregation)},
\]
whereas InfoNCE applies the logarithm per sample,
\[
    \mathbb{E}_{p(z_A)} \!\left[ \log \mathbb{E}_{p(z_B)} \!\left[e^{T(z_A, z_B)} \right] \right] \quad \text{(local aggregation)}.
\]
Although this difference affects the aggregation structure, both objectives have the joint-minus-marginal structure to which the proposition in \cref{sec:section_3.1} applies, under the same assumptions, and so support approximate monotonic improvement under alternating optimization.

\subsubsection{JSD}

Similarly, the JSD bound can be expressed as
\[
I_{\mathrm{JSD}}(Z_A; Z_B) =
\mathbb{E}_{p(z_A, z_B)}\!\big[-\log\!\big(1 + e^{-T(z_A, z_B)}\big)\big]
- \mathbb{E}_{p(z_A)p(z_B)}\!\big[\log\!\big(1 + e^{T(z_A, z_B)}\big)\big].
\]
This form corresponds to a binary classification objective, distinguishing samples from the joint distribution versus the product of marginals. As with DV and InfoNCE, it has a \textit{joint term minus marginal term} structure, but instead of a log-sum-exp aggregation, it applies the softplus nonlinearity independently to each sample.

All three objectives (DV, InfoNCE, JSD) have this structure, so the proposition in \cref{sec:section_3.1} applies to each under the same assumptions.

\subsection{Moment-based surrogates for the DV marginal term}
\label{app:moment_surrogates_for_DV_marginal}

We present the complete mathematical derivation referenced in \cref{sec:section_3.4} showing that the regularizers in Barlow Twins correspond to a second-order Taylor expansion (cumulant expansion) of the DV bound's marginal term.

\subsubsection*{DV bound}

To ground our approximation, we recall the DV bound (\cref{eq:dv_bound}):
\[
    I(Z_A; Z_B) \ge I_{\mathrm{DV}}(Z_A; Z_B) = \sup_{T \in \mathcal{F}} \left\{ \underbrace{\mathbb{E}_{p(z_A, z_B)}[T(z_A, z_B)]}_{\text{Joint term}} - \underbrace{\log \mathbb{E}_{p(z_A)p(z_B)}\!\left[e^{T(z_A, z_B)} \right]}_{\text{Marginal term}} \right\},
\]
where \(T \in \mathcal{F}\) is a critic function, chosen from a sufficiently expressive function class \(\mathcal{F}\).

\subsubsection*{CGF and Taylor expansion}

Let \(T\) be any bounded critic with
\[
  T(x,y)\in[a,b]\quad\text{for all }x,y,
\]
and define its CGF as
\begin{equation}
    \label{eq:cgf}
    K(s) = \log \mathbb{E}[e^{sT}].
\end{equation}
Because \(T\) is bounded, \(K\) is infinitely differentiable on \([0,1]\). Its \(n\)-th derivative at zero yields the \(n\)-th cumulant,
\[
    \kappa_n = K^{(n)}(0),
\]
and in particular,
\[
    K'(0)=\mathbb{E}[T],
    \quad
    K''(0)=\mathrm{Var}(T).
\]
By Taylor's theorem about \(s=0\), for \(s\in[0,1]\),
\begin{equation}
    \label{eq:taylor_series}
    K(s) = s\,K'(0) + \tfrac12 s^2\,K''(0) + R_2(s),
\end{equation}
where \(R_2(s) = \tfrac16 s^3 K^{(3)}(s^*)\) for some \(s^*\in(0,s)\).

Since \(T \in [a,b]\), all derivatives of \(K(s)\) are bounded on \([0,1]\). In particular, evaluating \cref{eq:taylor_series} at \(s = 1\) yields
\[
  \bigl|R_2(1)\bigr|
  \le \frac{1}{6}\max_{s\in[0,1]}\bigl|K^{(3)}(s)\bigr|
  = \mathcal{O}(1).
\]
The residual is therefore bounded, and the second-order approximation
\begin{equation}
    \label{eq:second_order}
    \log \mathbb{E}[e^T] = K(1)
    \approx \mathbb{E}[T] + \tfrac12 \mathrm{Var}(T)
\end{equation}
holds with controlled error.

\subsubsection*{Surrogate loss via mean--variance}

Substituting \cref{eq:second_order} into the DV bound (\cref{eq:dv_bound}) gives the moment-based surrogate
\begin{align}
\label{taylor_dv_bound}
    I_{\mathrm{DV}}(Z_A; Z_B)
    &\;\approx\; I_{\mathrm{Taylor\text{-}DV}}(Z_A; Z_B) \nonumber \\
    &= \mathbb{E}_{p(z_A, z_B)}[T(z_A, z_B)]
       - \left\{ \mathbb{E}_{p(z_A)p(z_B)}[T(z_A, z_B)]
       + \tfrac12\,\mathrm{Var}_{p(z_A)p(z_B)}\bigl[T(z_A, z_B)\bigr] \right\}.
\end{align}
We may construct a tractable loss as in \cref{eq:taylor_dv_loss}
\[
    \mathcal{L}_{\mathrm{Taylor\text{-}DV}} = -
    \underbrace{\mathbb{E}_{p(z_A,z_B)}[T(z_A,z_B)]}_{\text{Joint term}}
    +
    \underbrace{\mathbb{E}_{p(z_A)p(z_B)}[T(z_A,z_B)]}_{\text{Marginal mean term}}
    +
    \underbrace{\mathrm{Var}_{p(z_A)p(z_B)}[T(z_A,z_B)]}_{\text{Marginal variance term}}\,,
\]
where the \(\tfrac{1}{2}\) coefficient on the variance term is omitted for simplicity, as it can be absorbed into a tuning hyperparameter.

\subsubsection*{Barlow Twins as a mean--variance surrogate}

We use \(z^A_i\) and \(z^B_i\) for the \(i\)-th coordinate of views \(A\) and \(B\), respectively. Starting from \cref{eq:barlow_twins_dot},
\[
    X_i
    = z^A_i\,z^B_i,
    \quad
    T_{\cos}(z^A, z^B)
    = \sum_{i=1}^d X_i
    = z^A \cdot z^B.
\]
By the variance-of-a-sum identity,
\begin{equation}
      \mathrm{Var}\bigl[T(z^A,z^B)\bigr]
      = \mathrm{Var}\!\left(\sum_{i=1}^d X_i\right)
      = \sum_{i,j=1}^d \mathrm{Cov}(X_i,\,X_j)
      = \sum_{i,j=1}^d \mathrm{Cov}(z^A_i z^B_i,\;z^A_j z^B_j).
\end{equation}

Barlow Twins reduces the surrogate in \cref{eq:taylor_dv_loss} to an alignment term and a tractable approximation of the marginal variance by applying batch normalization, ensuring
\[
    \mathbb{E}_{p(z_A)}[z^A_i] = \mathbb{E}_{p(z_B)}[z^B_i] \approx 0 \quad \Rightarrow \quad \mathbb{E}_{p(z_A)p(z_B)}[T(z_A, z_B)] \approx 0.
\]

The regularization terms in Barlow Twins are constructed using batch-level statistics, specifically the \emph{cross-correlation matrix} between features across the two views,
\[
C_{ij} = \frac{1}{N} \sum_{n=1}^N z^A_{n,i} \cdot z^B_{n,j},
\]
where \(z^A_{n,i}\) and \(z^B_{n,j}\) denote the \(i\)-th and \(j\)-th features of the \(n\)-th sample from each view in a batch of size \(N\). The diagonal elements \(C_{ii}\) appear in the alignment term of the loss, encouraging each feature to match across views, while the off-diagonal elements \(C_{ij}\) for \(i \ne j\) are penalized to reduce redundancy.

To connect this to the variance term in \cref{eq:taylor_dv_loss}, we compute the covariance under independent sampling. Since \(z^A\) and \(z^B\) are drawn independently and batch normalization gives \(\mathbb{E}[z^A_i] = \mathbb{E}[z^B_i] \approx 0\), each product term has approximately zero mean, and grouping the factors by view gives
\begin{equation}
    \mathrm{Cov}(z^A_i z^B_i,\; z^A_j z^B_j)
    = \mathbb{E}\big[z^A_i z^A_j\big]\,\mathbb{E}\big[z^B_i z^B_j\big].
\end{equation}
Summing over coordinates and using \(\mathbb{E}[(z^A_i)^2] = \mathbb{E}[(z^B_i)^2] = 1\) under batch normalization yields
\begin{equation}
    \mathrm{Var}_{p(z_A)p(z_B)}\!\left[\sum_{i=1}^d z^A_i z^B_i\right]
    = d + \sum_{i \ne j} \mathbb{E}\big[z^A_i z^A_j\big]\,\mathbb{E}\big[z^B_i z^B_j\big].
\end{equation}
The constant \(d\) does not depend on the encoder parameters and contributes no gradient. The remaining approximation is that, once the alignment term has drawn the two views together, these within-view moments track the cross-correlation matrix, \(\mathbb{E}[z^A_i z^A_j] \approx \mathbb{E}[z^B_i z^B_j] \approx C_{ij}\), so that
\[
    \mathrm{Var}\!\left[\sum_i z^A_i z^B_i\right] \approx d + \sum_{i \ne j} C_{ij}^2,
\]
where the constant can be dropped from the loss.

Putting everything together, the Taylor--DV surrogate yields \cref{eq:bt_taylor_dv_surrogate},
\[
    \mathcal{L}_{\mathrm{Taylor\text{-}DV}} \approx -\sum_{i=1}^d C_{ii} + \sum_{i \ne j} C_{ij}^2,
\]
which matches the structure of the empirical Barlow Twins loss: an alignment term encouraging the diagonal of the cross-correlation matrix to approach \(1\), and a decorrelation term penalizing off-diagonal elements.

We validate the alignment approximation numerically. We draw two views of a shared Gaussian latent with within-view correlation \(\rho\), add independent per-feature noise to each view, standardize each feature as batch normalization does, and compare the exact marginal-product variance of the dot product against \(d + \sum_{i\ne j} C_{ij}^2\) for \(d = 32\) features. As shown in \cref{tab:bt_surrogate_check}, the surrogate tracks the variance to within one percent even at \(\rho = 0.8\), where the features are strongly correlated within each view.

\begin{table}[H]
\centering
\caption{Numerical check of the variance surrogate under increasing within-view correlation \(\rho\) (\(d = 32\), \(\sigma = 0.3\), Monte Carlo with \(2 \times 10^5\) samples).}
\label{tab:bt_surrogate_check}
\begin{tabular}{@{}lccc@{}}
\toprule
 & \(\mathrm{Var}_{p(z_A)p(z_B)}[T]\) & \(d + \sum_{i\ne j} C_{ij}^2\) & Relative error \\
\midrule
\(\rho = 0.0\) & 32.1 & 32.0 & 0.4\% \\
\(\rho = 0.2\) & 65.2 & 65.4 & 0.2\% \\
\(\rho = 0.4\) & 167.5 & 166.5 & 0.6\% \\
\(\rho = 0.6\) & 334.4 & 332.4 & 0.6\% \\
\(\rho = 0.8\) & 568.1 & 565.6 & 0.5\% \\
\bottomrule
\end{tabular}
\end{table}

The approximation relies on the alignment term having drawn the two views together, so we also vary the strength of the augmentation noise \(\sigma\), which controls how much content the two views share. Alignment is measured as the mean of the diagonal \(C_{ii}\). As shown in \cref{tab:bt_surrogate_alignment}, the agreement holds as alignment falls from 0.99 to 0.31, well below the values reached in practice.

\begin{table}[H]
\centering
\caption{Numerical check of the variance surrogate under weakening alignment (\(d = 32\), \(\rho = 0.2\), Monte Carlo with \(2 \times 10^5\) samples). Alignment is the mean of the diagonal \(C_{ii}\), and \(\sigma\) is the standard deviation of the added view noise.}
\label{tab:bt_surrogate_alignment}
\begin{tabular}{@{}lcccc@{}}
\toprule
 & Alignment & \(\mathrm{Var}_{p(z_A)p(z_B)}[T]\) & \(d + \sum_{i\ne j} C_{ij}^2\) & Relative error \\
\midrule
\(\sigma = 0.1\) & 0.99 & 70.6 & 71.1 & 0.6\% \\
\(\sigma = 0.3\) & 0.92 & 65.4 & 65.1 & 0.5\% \\
\(\sigma = 0.6\) & 0.74 & 53.1 & 53.5 & 0.8\% \\
\(\sigma = 1.0\) & 0.50 & 41.9 & 42.0 & 0.1\% \\
\(\sigma = 1.5\) & 0.31 & 35.8 & 35.8 & 0.1\% \\
\bottomrule
\end{tabular}
\end{table}

\subsection{Input informativeness}
\label{app:input_informativeness}

While the SDMI and JMI frameworks increase \(I(z_E; z_M)\), their effectiveness depends on how this relates to the input \(x\). We formalize this intuition with the following proposition.

\paragraph{Proposition (view-MI is bounded by input informativeness).}
Assume that, conditional on the input \(x\), the two augmented views \(x_1, x_2\) are
drawn independently from the augmentation distribution \(A(x)\), so that
\(z^{(1)} \perp\!\!\!\perp z^{(2)} \mid x\). Then
\begin{equation}
    I(z^{(1)};z^{(2)})
    \;\leq\;
    \min\bigl\{\, I(x;z^{(1)}),\; I(x;z^{(2)}) \,\bigr\}.
\end{equation}

\paragraph{Proof.}
Under the stated assumption, \(z^{(1)} \leftrightarrow x \leftrightarrow z^{(2)}\)
forms a Markov chain. The Data Processing Inequality then gives
\[
    I(z^{(1)}; z^{(2)}) \;\leq\; I(x; z^{(1)}),
\]
and by symmetry \(I(z^{(1)}; z^{(2)}) \leq I(x; z^{(2)})\). Combining the two
inequalities yields
\[
    I(z^{(1)}; z^{(2)}) \;\leq\; \min\bigl\{\, I(x; z^{(1)}),\; I(x; z^{(2)}) \,\bigr\}.
    \qed
\]

\section{Additional experimental results}
\label{app:additional_experimental_results}

Supplementary experimental results and analyses support the findings presented in \cref{sec:experiment}, including controlled experiments on synthetic data and additional ablation studies.

\subsection{Controlled experiment}
\label{sec:toy_experiment}

\subsubsection{SDMI deterministic full-batch updates}
\label{app:sdmi_full_batch}

\paragraph{Nearest-neighbor angle statistics.}
To complement \cref{fig:toy_center_movement}, we report rotation-invariant NN angle statistics for the final cluster embeddings in \cref{tab:nn_gap}. In our setup, on a 3D unit sphere, the ideal separation for five clusters corresponds to a NN gap of \(\approx 90^\circ\) (Thomson optimum).

\begin{table}[h]
  \centering
  \caption{NN angle gaps at convergence.}
  \label{tab:nn_gap}
  \begin{tabular}{@{}lcccc@{}}
    \toprule
    \textbf{Model} & \textbf{Mean NN Gap ($^\circ$)} & \textbf{Min NN Gap ($^\circ$)} & \textbf{Max NN Gap ($^\circ$)} & \textbf{SD ($^\circ$)} \\
    \midrule
    SDMI prototype (Encoder $f_\theta$) & \textbf{74.47} & \textbf{72.78} & \textbf{77.60} & 1.76 \\
    SDMI prototype (Encoder $g_\xi$)    & 73.77          & 72.16          & 75.06          & \textbf{1.32} \\
    SimSiam                             & 40.62          & 28.04          & 71.86          & 16.19 \\
    MoCo                                & 49.64          & 47.90          & 54.07          & 2.29 \\
    BYOL                                & 37.35          & 31.26          & 45.34          & 5.42 \\
    \midrule
    JMI prototype                       & \textbf{88.26} & \textbf{85.49} & \textbf{94.03} & \textbf{3.36} \\
    SimCLR                              & 65.80          & 61.95          & 70.60          & 3.38 \\
    Barlow Twins                        & 81.62          & 64.46          & 102.53         & 15.55 \\
    VICReg                              & 82.15          & 69.28          & 100.95         & 14.54 \\
    \bottomrule
  \end{tabular}
\end{table}

\paragraph{Cluster dynamics.}

\begin{figure}[t]
  \centering
  \includegraphics[width=1\textwidth]{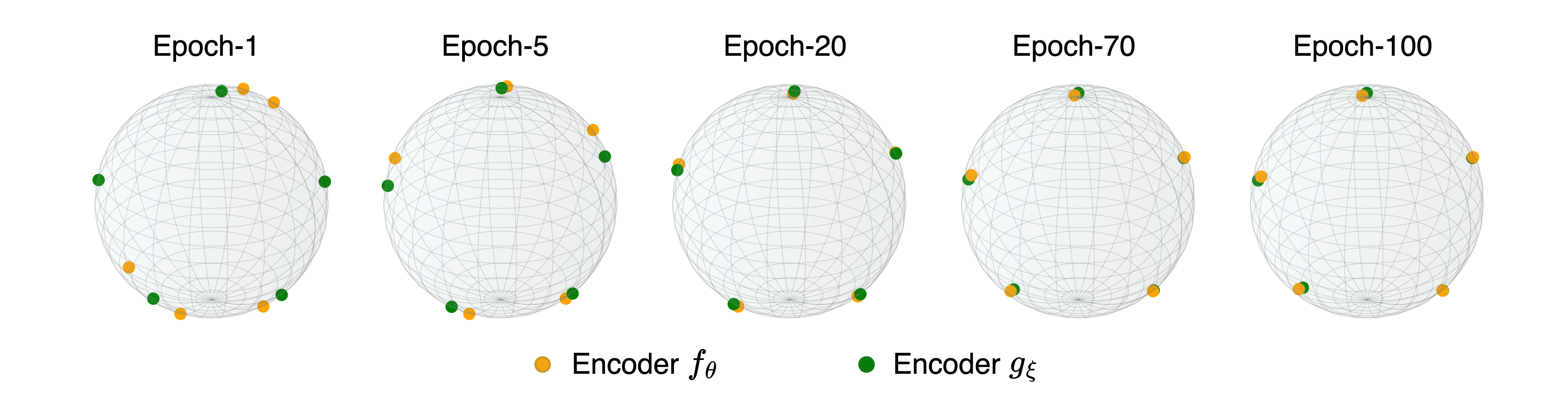}
  \caption{Cluster centers on the unit sphere, showing how SDMI prototype encoders progressively separate the five Gaussian clusters across training.}
  \label{fig:toy_cluster_separation}
\end{figure}

In \cref{fig:toy_cluster_separation}, we plot the embeddings of each centroid under both \(f_\theta\) and \(g_\xi\) encoders at various training iterations (epochs \(1, 5, 20, 70, 100\)). As training progresses, the two views of each cluster increasingly align with one another, while the embeddings across different clusters become progressively more separated, indicating that the independently updated encoders learn both consistent and discriminative representations.

\subsubsection{SDMI stochastic mini-batch updates}
\label{app:sdmi_stochastic}

\begin{figure}[H]
  \centering
  \includegraphics[width=1\textwidth]{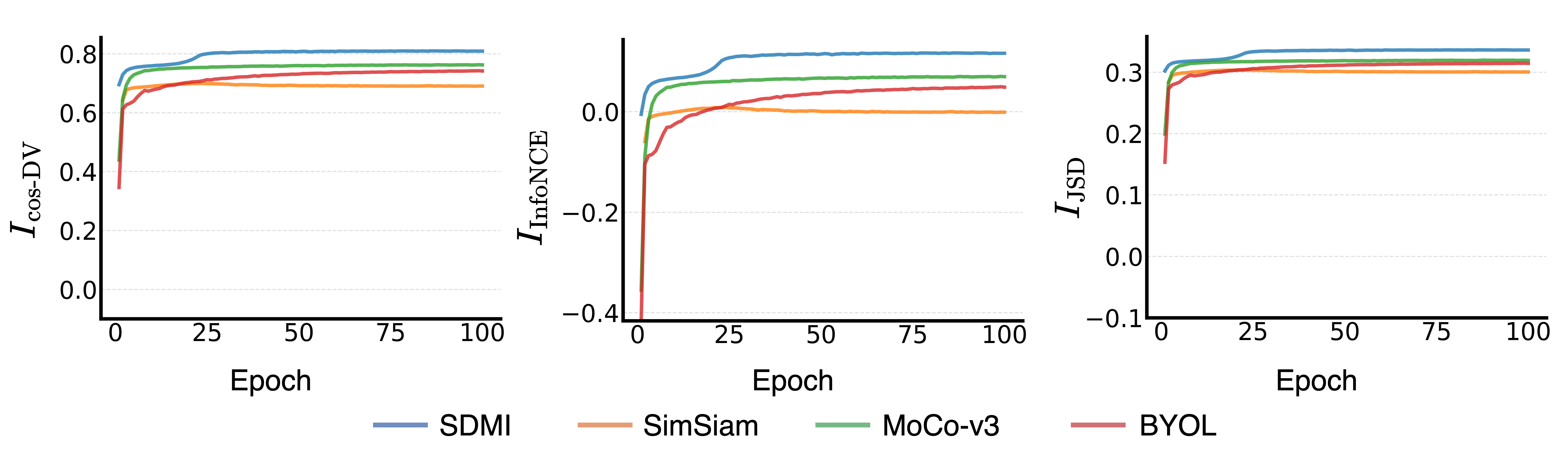}
  \caption{Estimated MI under mini-batch SGD. All methods continue to show monotonic MI growth across three estimators (\(I_{\mathrm{cos\text{-}DV}}\), \(I_{\mathrm{InfoNCE}}\), \(I_{\mathrm{JSD}}\)).}
  \label{fig:sdmi_sgd_mi}
\end{figure}

To examine whether the SDMI prototype continues to maximize MI under stochastic optimization, we rerun the toy Gaussian mixture experiment using a training batch size of 500 and standard SGD (learning rate 0.5, cosine annealing schedule, 100 epochs). \Cref{fig:sdmi_sgd_mi} shows the estimated \(I_{\mathrm{cos\text{-}DV}}\), \(I_{\mathrm{InfoNCE}}\), and \(I_{\mathrm{JSD}}\) curves for SDMI and the three baseline methods. All approaches continue to show approximately monotonic MI growth despite the use of mini-batches.

\subsection{Effect of target proximity and update speed in EMA-based self-distillation}
\label{app:ema_closeness}

\begin{figure}[t]
  \centering
  \includegraphics[width=\textwidth]{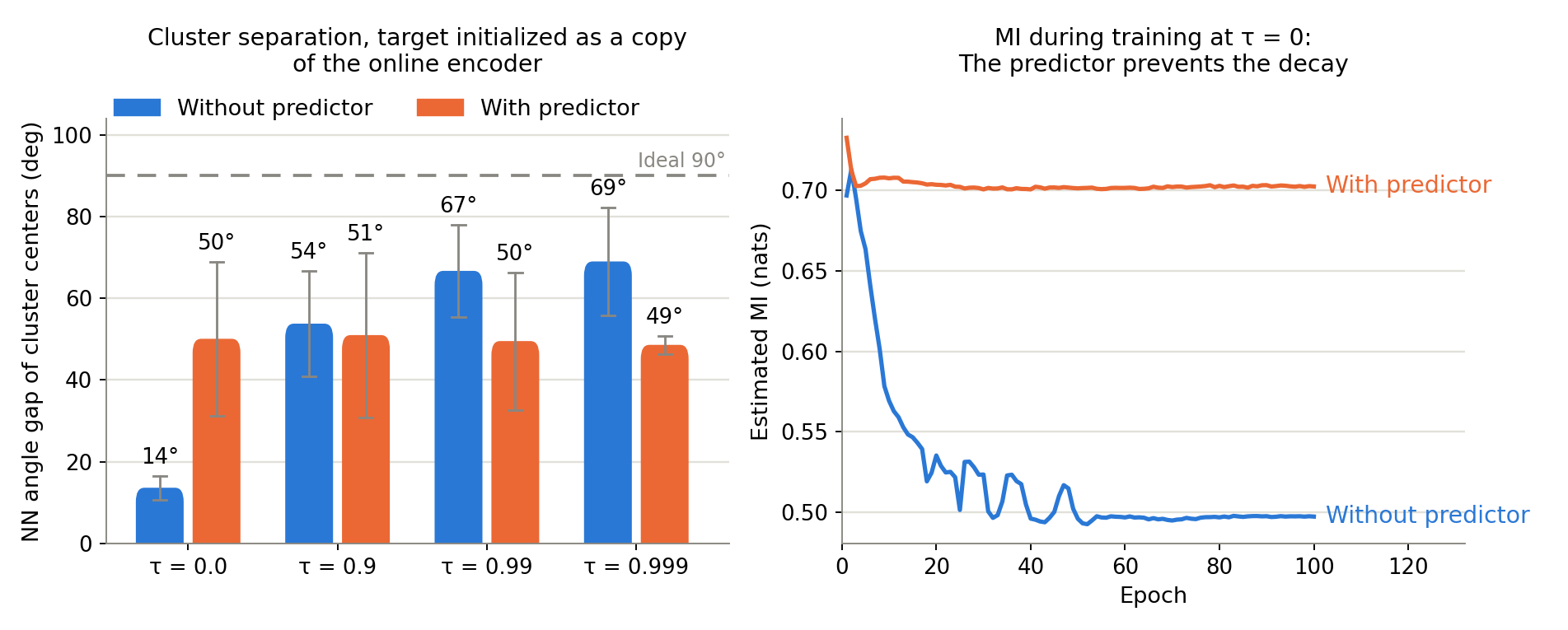}
  \caption{EMA-closeness ablation, target initialized as a copy of the online encoder. Left, NN angle gap of cluster centers (mean $\pm$ standard deviation, three seeds). Right, estimated MI (cos-DV) at $\tau = 0$, where the predictor prevents the decay.}
  \label{fig:ema_ablation_summary}
\end{figure}

\begin{figure}[t]
  \centering
  \includegraphics[width=\textwidth]{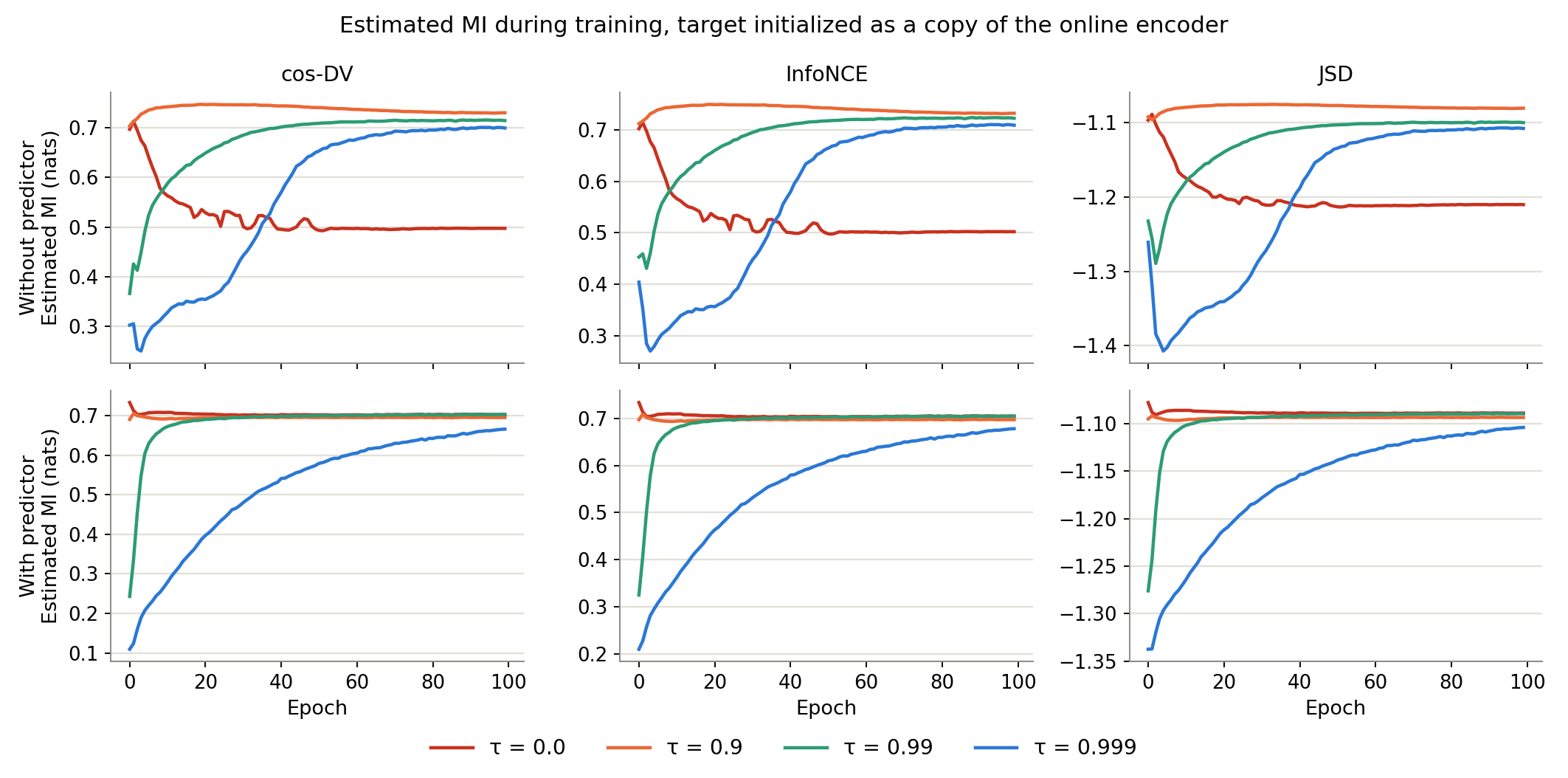}
  \caption{Estimated MI during training under the cos-DV, InfoNCE, and JSD bounds (left to right), without (top) and with (bottom) a predictor. All three estimators agree.}
  \label{fig:ema_ablation_mi}
\end{figure}

\begin{figure}[t]
  \centering
  \includegraphics[width=\textwidth]{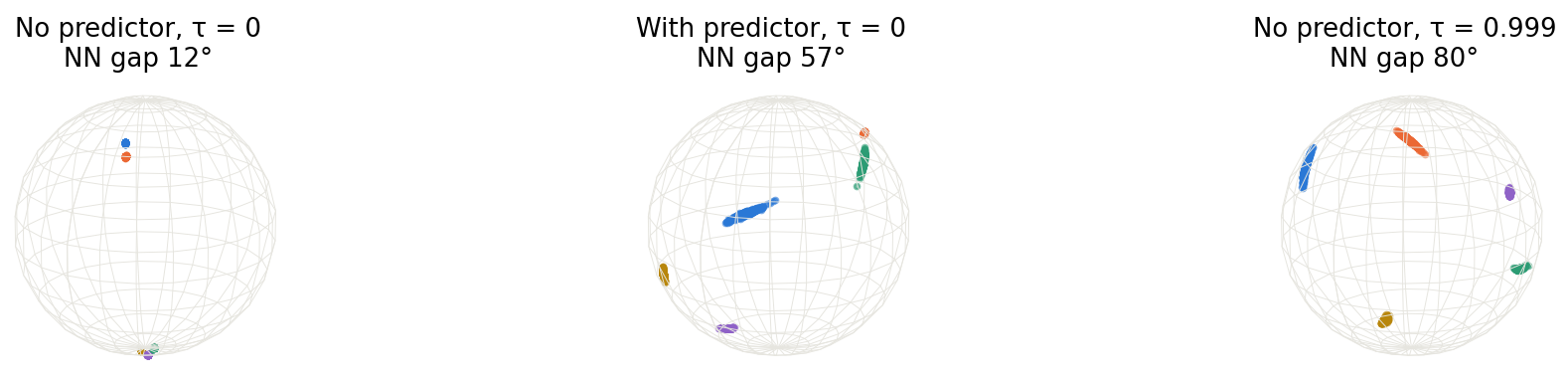}
  \caption{Final embeddings on the unit sphere for three representative configurations. Clusters collapse without a predictor at $\tau = 0$, remain separated with a predictor, and separate further with a slow target.}
  \label{fig:ema_ablation_embeddings}
\end{figure}

A natural failure mode for EMA-based self-distillation arises when the target encoder stays too close to the online encoder, through initialization or through an aggressive momentum coefficient in \cref{eq:byol_m_step}. We study this setting on the controlled Gaussian setup of \cref{sec:experimental_setup_toy} with a BYOL-style model, initializing the target as an exact copy of the online encoder and sweeping the momentum coefficient $\tau \in \{0, 0.9, 0.99, 0.999\}$, with and without a predictor, over three seeds; $\tau = 0$ copies the online encoder into the target at every step. \Cref{fig:ema_ablation_summary} reports the NN angle gap of the cluster centers (\cref{sec:results}) together with the estimated MI during training, and \cref{fig:ema_ablation_mi} reports all three estimators.

Without a predictor, the fastest target ($\tau = 0$) collapses the clusters (NN gap $13.6^\circ \pm 3.0$) and the estimated MI decays during training, whereas adding a predictor restores separation ($50.1^\circ \pm 18.9$) and holds MI stable, so the predictor's surrogate role does not require separation between the branches. A slowly drifting target alone is also sufficient ($69.0^\circ \pm 13.3$ at $\tau = 0.999$ without a predictor), consistent with the slowly varying marginal statistics discussed in \cref{sec:section_3.1}, although combining the predictor with a slow target does not improve over the slow target alone at this scale. \Cref{fig:ema_ablation_embeddings} visualizes the learned embeddings in the three representative configurations.

\section{Implementation details}
\label{app:implementation_details}

\subsection{Compute resources}

All experiments were conducted on two servers equipped with an NVIDIA RTX 5090 GPU and an NVIDIA RTX 4080 GPU. The complete set of experiments, including hyperparameter sweeps and additional experiments not included in the paper, took approximately 6 weeks of wall-clock time. For the SDMI prototype, GPU memory (VRAM) usage was around 15 GB, while all other methods required approximately 7 GB.

\subsection{Real data benchmark}

\paragraph{Hyperparameter sweep 1.}
We conduct a grid search for each model using a ResNet-18 encoder backbone on the CIFAR10/100 datasets. The hyperparameters explored in the initial sweep are summarized in \cref{tab:hyperparams}. Models requiring a predictor network (e.g., SimSiam, BYOL, and MoCo-v3) use a fixed 2-layer predictor. For methods that incorporate a temperature parameter (e.g., SDMI, MoCo-v3, JMI, and SimCLR), an additional dimension is included in the search space. Momentum-based models such as BYOL and MoCo-v3 use a fixed momentum coefficient of 0.996 for the target encoder. The total number of configurations evaluated per model is shown in \cref{tab:configs}.

\paragraph{Hyperparameter sweep 2.}
Based on the results from sweep 1, we perform a secondary evaluation for each model on each dataset using 3 projection layers. The top-performing configuration for each model (shown in \cref{tab:optimal_hyperparameters}) is then selected and used to train that model for 1000 epochs.

\begin{table}[h]
\centering
\caption{Sweep 1: hyperparameter settings and search space used in our grid search.}
\label{tab:hyperparams}
\begin{tabular}{@{}lccc@{}}
\toprule
\textbf{Parameter} & \textbf{Values} & \textbf{Applies to} & \textbf{Fixed/Varied} \\
\midrule
Encoder Backbone     & ResNet-18                 & All models/datasets             & Fixed  \\
Batch Size           & 512                       & All models/datasets             & Fixed  \\
Projection Layers    & 2                         & All models/datasets             & Fixed  \\
Prediction Layers    & 2                         & All models/datasets             & Fixed  \\
Prediction Dimension & 256                       & All models/datasets             & Fixed  \\
Epochs               & 300                       & All models/datasets             & Fixed  \\
Feature Dimension    & 2048                      & All models/datasets             & Fixed  \\
Momentum Coefficient & 0.996                     & BYOL, MoCo-v3                   & Fixed  \\
Seed                 & 7349                      & All models/datasets             & Fixed  \\
Learning Rate        & \{0.01, 0.03, 0.05\}      & All models (cosine decay)       & Varied \\
Weight Decay         & \{0.0001, 0.0005\}        & All models                      & Varied \\
Temperature          & \{0.05, 0.07, 0.1\}       & SDMI, JMI, MoCo-v3, SimCLR      & Varied \\
Projection Dimension & \{128, 256\}              & All models                      & Varied \\
\bottomrule
\end{tabular}
\end{table}

\begin{table}[h]
\centering
\caption{Grid search configuration counts per model for sweep 1. Configurations are counted per dataset unless otherwise noted.}
\label{tab:configs}
\begin{tabular}{@{}lccc@{}}
\toprule
\textbf{Model} & \textbf{Dataset(s)} & \textbf{Temperature Used} & \textbf{Total Configurations} \\
\midrule
SDMI         & CIFAR10, CIFAR100 & Yes & $3 \times 2 \times 3 \times 2 = 36$ \\
SimSiam      & CIFAR10, CIFAR100 & No  & $3 \times 2 \times 2 = 12$ \\
BYOL         & CIFAR10, CIFAR100 & No  & $3 \times 2 \times 2 = 12$ \\
MoCo-v3      & CIFAR10, CIFAR100 & Yes & $3 \times 2 \times 3 \times 2 = 36$ \\
JMI          & CIFAR10, CIFAR100 & Yes & $3 \times 2 \times 3 \times 2 = 36$ \\
SimCLR       & CIFAR10, CIFAR100 & Yes & $3 \times 2 \times 3 \times 2 = 36$ \\
Barlow Twins & CIFAR10, CIFAR100 & No  & $3 \times 2 \times 2 = 12$ \\
VICReg       & CIFAR10, CIFAR100 & No  & $3 \times 2 \times 2 = 12$ \\
\bottomrule
\end{tabular}
\end{table}

\begin{table}[h]
\centering
\caption{Optimal hyperparameters selected from sweep 2 for each model and dataset. LR = learning rate, WD = weight decay, Temp. = temperature. All models use a ResNet-18 encoder. For VICReg, we fix the similarity, variance, and covariance loss coefficients to 25.0, 25.0, and 1.0, respectively, and set a small numerical stability term $\epsilon = 10^{-4}$. These values remain constant across all runs.}
\label{tab:optimal_hyperparameters}
\begin{tabular}{@{}llcccccc@{}}
\toprule
\textbf{Model} & \textbf{Dataset} & \textbf{LR} & \textbf{WD} & \textbf{Temp.} & \textbf{Proj. Dim} & \textbf{Proj. Layers} & \textbf{Predictor} \\
\midrule
SDMI         & CIFAR10  & 0.05 & 0.0005 & 0.1  & 256 & 3 & No  \\
SDMI         & CIFAR100 & 0.03 & 0.0005 & 0.1  & 256 & 2 & No  \\
SimSiam      & CIFAR10  & 0.05 & 0.0005 & --   & 256 & 3 & Yes \\
SimSiam      & CIFAR100 & 0.05 & 0.0005 & --   & 256 & 3 & Yes \\
BYOL         & CIFAR10  & 0.05 & 0.0005 & --   & 256 & 3 & Yes \\
BYOL         & CIFAR100 & 0.05 & 0.0005 & --   & 256 & 2 & Yes \\
MoCo-v3      & CIFAR10  & 0.05 & 0.0005 & 0.1  & 256 & 2 & Yes \\
MoCo-v3      & CIFAR100 & 0.05 & 0.0001 & 0.1  & 256 & 2 & Yes \\
JMI          & CIFAR10  & 0.05 & 0.0005 & 0.1  & 128 & 2 & No  \\
JMI          & CIFAR100 & 0.03 & 0.0005 & 0.1  & 256 & 2 & No  \\
SimCLR       & CIFAR10  & 0.05 & 0.0005 & 0.1  & 128 & 2 & No  \\
SimCLR       & CIFAR100 & 0.03 & 0.0001 & 0.1  & 128 & 2 & No  \\
Barlow Twins & CIFAR10  & 0.05 & 0.0005 & --   & 256 & 2 & No  \\
Barlow Twins & CIFAR100 & 0.05 & 0.0005 & --   & 256 & 2 & No  \\
VICReg       & CIFAR10  & 0.05 & 0.0005 & --   & 256 & 3 & No  \\
VICReg       & CIFAR100 & 0.03 & 0.0005 & --   & 256 & 3 & No  \\
\bottomrule
\end{tabular}
\end{table}

\subsection{ImageNet100 and TinyImageNet training configuration}
\label{app:imagenet100_tinyimagenet}

For both the ImageNet100 and TinyImageNet datasets, we adopt a uniform configuration across all models, with only a few dataset-specific adjustments. The settings are summarized in \cref{tab:imagenet100_tinyimagenet_hyperparams}.

\begin{table}[H]
\centering
\caption{Hyperparameter settings used for all models trained on ImageNet100 and TinyImageNet.}
\label{tab:imagenet100_tinyimagenet_hyperparams}
\begin{tabular}{@{}lcc@{}}
\toprule
\textbf{Parameter} & \textbf{ImageNet100} & \textbf{TinyImageNet} \\
\midrule
Encoder Backbone       & ResNet-50      & ResNet-18    \\
Epochs                 & 800            & 1000         \\
Warmup Epochs          & 10             & 5            \\
Batch Size             & 256            & 512          \\
Initial Learning Rate  & 0.05           & 0.05         \\
Learning Rate Schedule & Cosine decay   & Cosine decay \\
Weight Decay           & 0.0001         & 0.0005       \\
Projection Layers      & 3              & 3            \\
Projection Dimension   & 256            & 256          \\
Feature Dimension      & 512            & 512          \\
\bottomrule
\end{tabular}
\end{table}

\paragraph{Note.}
For the SDMI model on ImageNet100, the batch size is reduced to 64 (instead of 256) because of the increased memory requirements from the two-encoder setup.

\section{Method classification under the SDMI/JMI taxonomy}
\label{app:method_classification}

\Cref{tab:method_classification} classifies representative SSRL methods under the SDMI/JMI framework developed in \cref{sec:method}, categorizing each method by its update rule (EM-style alternating vs.\ joint) and by whether it optimizes an explicit MI bound or an MI surrogate.

\begin{table}[H]
\centering
\caption{Representative SSRL methods and their classification under the SDMI/JMI taxonomy, with objective types.}
\label{tab:method_classification}
\renewcommand{\arraystretch}{1.2}
\begin{tabular}{lccc}
\toprule
\textbf{Method} & \textbf{EM/Joint} & \textbf{Objective Type} & \textbf{Paradigm} \\
\midrule
BYOL \citep{grill2020bootstrap}                                                   & EM    & MI Surrogate & SDMI \\
SimSiam \citep{chen2020exploring}                                                 & EM    & MI Surrogate & SDMI \\
MoCo \citep{he2020momentum, chen2020improved, chen2021empirical}                  & EM    & Explicit MI  & SDMI \\
DINO \citep{caron2021emerging}                                                    & EM    & MI Surrogate & SDMI \\
\midrule
SimCLR \citep{chen2020simple}                                                     & Joint & Explicit MI  & JMI \\
Barlow Twins \citep{zbontar2021barlow}                                            & Joint & MI Surrogate & JMI \\
VICReg \citep{bardes2021vicreg}                                                   & Joint & MI Surrogate & JMI \\
W-MSE \citep{ermolov2021whitening}                                                & Joint & MI Surrogate & JMI \\
SwAV \citep{caron2020unsupervised}                                                & Joint & MI Surrogate & JMI \\
BGRL \citep{thakoor2021large}                                                     & Joint & MI Surrogate & JMI \\
\bottomrule
\end{tabular}
\end{table}

\section{Algorithms}
\label{app:algorithms}

Detailed algorithmic descriptions for the canonical SDMI and JMI prototypes introduced in \cref{sec:canonical_forms} and illustrated in \cref{fig:sdmi_jmi} are provided below. Following common SSRL practice~\citep{chen2020simple,chen2021empirical,grill2020bootstrap,chen2020exploring}, we adopt a symmetric loss by computing the objective over both view orderings.

\subsection{SDMI canonical form training procedure}
In all of our experiments, the E-step and M-step each complete a full pass over the dataset before alternating, regardless of dataset size. The block-coordinate interpretation does not depend on this granularity, and alternation could equally be performed at the mini-batch level; we use epoch-level alternation because it keeps the frozen branch fixed over many updates, yielding stable targets and a clean correspondence with the EM procedure in \cref{sec:EM_style_MI_maximization}.

\begin{algorithm}[H]
\SetAlgoLined
\KwIn{Unlabeled dataset \(\mathcal{D}\), encoders \(f_\theta\), \(g_\xi\), temperature \(\tau\), number of epochs \(T\)}
\KwOut{Trained encoder parameters \(\theta\), \(\xi\)}
\For{\(t = 1\) \KwTo \(T\)}{
    \tcp*[h]{E-Step: Update \(f_\theta\), freeze \(g_\xi\)}
    \ForEach{minibatch \((X_1, X_2) \sim \mathcal{D}\)}{
        \(Z_E^{(1)} \leftarrow f_\theta(X_1)\),\quad \(Z_E^{(2)} \leftarrow f_\theta(X_2)\) \\
        \(Z_M^{(1)} \leftarrow g_\xi(X_1)\),\quad \(Z_M^{(2)} \leftarrow g_\xi(X_2)\) \\
        \(\hat{Z}_M^{(1)} \leftarrow \text{SG}(Z_M^{(1)}),\quad \hat{Z}_M^{(2)} \leftarrow \text{SG}(Z_M^{(2)})\) \\
        \(\mathcal{L}_E \leftarrow \frac{1}{2} \left[ \mathrm{DV}(Z_E^{(1)}, \hat{Z}_M^{(2)}; \tau) + \mathrm{DV}(Z_E^{(2)}, \hat{Z}_M^{(1)}; \tau) \right]\) \\
        Update \(\theta\) via gradient descent on \(\mathcal{L}_E\)
    }
    \tcp*[h]{M-Step: Update \(g_\xi\), freeze \(f_\theta\)}
    \ForEach{minibatch \((X_1, X_2) \sim \mathcal{D}\)}{
        \(Z_E^{(1)} \leftarrow f_\theta(X_1)\),\quad \(Z_E^{(2)} \leftarrow f_\theta(X_2)\) \\
        \(Z_M^{(1)} \leftarrow g_\xi(X_1)\),\quad \(Z_M^{(2)} \leftarrow g_\xi(X_2)\) \\
        \(\hat{Z}_E^{(1)} \leftarrow \text{SG}(Z_E^{(1)}),\quad \hat{Z}_E^{(2)} \leftarrow \text{SG}(Z_E^{(2)})\) \\
        \(\mathcal{L}_M \leftarrow \frac{1}{2} \left[ \mathrm{DV}(Z_M^{(1)}, \hat{Z}_E^{(2)}; \tau) + \mathrm{DV}(Z_M^{(2)}, \hat{Z}_E^{(1)}; \tau) \right]\) \\
        Update \(\xi\) via gradient descent on \(\mathcal{L}_M\)
    }
}
\caption{EM-style Training Procedure of the SDMI Prototype}
\label{alg:sdmi_train_loop}
\end{algorithm}

\subsection{JMI canonical form training procedure}

\begin{algorithm}[H]
\SetAlgoLined
\KwIn{Unlabeled dataset \(\mathcal{D}\), encoder \(f_\theta\), temperature \(\tau\), number of epochs \(T\)}
\KwOut{Trained encoder parameters \(\theta\)}
\For{\(t = 1\) \KwTo \(T\)}{
    \ForEach{minibatch \((X_1, X_2) \sim \mathcal{D}\)}{
        \(Z^{(1)} \leftarrow f_\theta(X_1)\),\quad \(Z^{(2)} \leftarrow f_\theta(X_2)\) \\
        \(\mathcal{L} \leftarrow \frac{1}{2} \left[ \mathrm{DV}(Z^{(1)}, Z^{(2)}; \tau) + \mathrm{DV}(Z^{(2)}, Z^{(1)}; \tau) \right]\) \\
        Update \(\theta\) via gradient descent on \(\mathcal{L}\)
    }
}
\caption{Joint Training Procedure of the JMI Prototype}
\label{alg:jmi_train_loop}
\end{algorithm}

\end{document}